%% file: paper.tex
\documentclass[10pt,twocolumn,letterpaper]{article}

\usepackage[pagenumbers]{cvpr} 
\usepackage[accsupp]{axessibility} 

\usepackage{times}
\usepackage{epsfig}
\usepackage{graphicx}
\usepackage{amsmath}
\usepackage{amssymb}
\usepackage{mathtools}
\usepackage{booktabs}
\usepackage{subcaption}
\usepackage{color}
\usepackage{dcolumn}
\usepackage{float}
\usepackage{wrapfig}
\usepackage[export]{adjustbox}

\makeatletter
\@namedef{ver@everyshi.sty}{}
\makeatother

\usepackage[dvipsnames]{xcolor}
\usepackage{tikz}
\usepackage{graphicx}
\usepackage{pgfplots}
\usetikzlibrary{arrows, arrows.meta, shapes, shapes.misc, calc, positioning, decorations.pathreplacing, fit, backgrounds}

\usepackage[pagebackref,breaklinks,colorlinks]{hyperref}

\usepackage[capitalize]{cleveref}
\crefname{section}{Sec.}{Secs.}
\Crefname{section}{Section}{Sections}
\Crefname{table}{Table}{Tables}
\crefname{table}{Tab.}{Tabs.}

\def\cvprPaperID{4399} 
\def\confName{CVPR}
\def\confYear{2022}

\begin{document}

\title{Towards Multimodal Depth Estimation from Light Fields}

\author{Titus Leistner, Radek Mackowiak, Lynton Ardizzone, Ullrich Köthe, Carsten Rother\\
Visual Learning Lab, Heidelberg University\\
{\tt\small first.last@iwr.uni-heidelberg.de}
}

\definecolor{my-gray}{HTML}{444444}
\definecolor{my-orange}{HTML}{e37238}
\definecolor{my-green}{HTML}{44b52d}
\definecolor{my-blue}{HTML}{1a92ce}
\definecolor{my-purple}{HTML}{b84ace}

\maketitle

\begin{abstract}
\input{sections/abstract}
\end{abstract}

\section{Introduction}
\input{sections/introduction}

\section{Related Work}
\input{sections/related_work}

\section{Method}
\input{sections/method}

\section{Experiments}
\label{sec:experiments}
\input{sections/experiments}

\section{Conclusion}
\input{sections/conclusion}

\textbf{Acknowledgement:}
\input{sections/acknowledgement}

{\small
\bibliographystyle{ieee_fullname}
\bibliography{refs}
}

\newpage
\twocolumn[{%
    \begin{@twocolumnfalse}
        \null
        \vskip .375in
        \begin{center}
          {\Large \bf Appendix for:\\Towards Multimodal Depth Estimation from Light Fields \par}
          \vspace*{24pt}
          {
          \large
          \lineskip .5em
          \begin{tabular}[t]{c}
            Titus Leistner, Radek Mackowiak, Lynton Ardizzone, Ullrich Köthe, Carsten Rother
            \\
            Visual Learning Lab, Heidelberg University
          \end{tabular}
          \par
          }
          \vskip .5em
          \vspace*{12pt}
        \end{center}
    \end{@twocolumnfalse}
}]
\appendix

\renewcommand\thefigure{\thesection.\arabic{figure}}
\renewcommand\theequation{\thesection.\arabic{equation}}
\renewcommand\thetable{\thesection.\arabic{table}}

\newcommand{\asection}[1] {
    \setcounter{figure}{0}
    \setcounter{equation}{0}
    \setcounter{table}{0}
    \section{#1}
}
\renewcommand{\textfraction}{0.0}
\restylefloat{figure}

\input{sections/appendix.tex}
\end{document}

%% file: sections/abstract.tex
Light field applications, especially light field rendering and depth estimation, developed rapidly in recent years.
While state-of-the-art light field rendering methods handle semi-transparent and reflective objects well,
depth estimation methods either ignore these cases altogether or only deliver a weak performance.
We argue that this is due current methods only considering a single ``true'' depth, even when multiple objects at different depths contributed to the color of a single pixel.
Based on the simple idea of outputting a posterior depth distribution instead of only a single estimate, we develop and explore several different deep-learning-based approaches to the problem.
Additionally, we contribute the first ``multimodal light field depth dataset'' that contains the depths of all objects which contribute to the color of a pixel.
This allows us to supervise the multimodal depth prediction and also validate all methods by measuring the KL divergence of the predicted posteriors.
With our thorough analysis and novel dataset, we aim to start a new line of depth estimation research that overcomes some of the long-standing limitations of this field. 

%% file: sections/introduction.tex
\begin{figure}[h]
    \centering
    \begin{subfigure}{0.495\textwidth}
        \centering
        \includegraphics[width=0.495\textwidth]{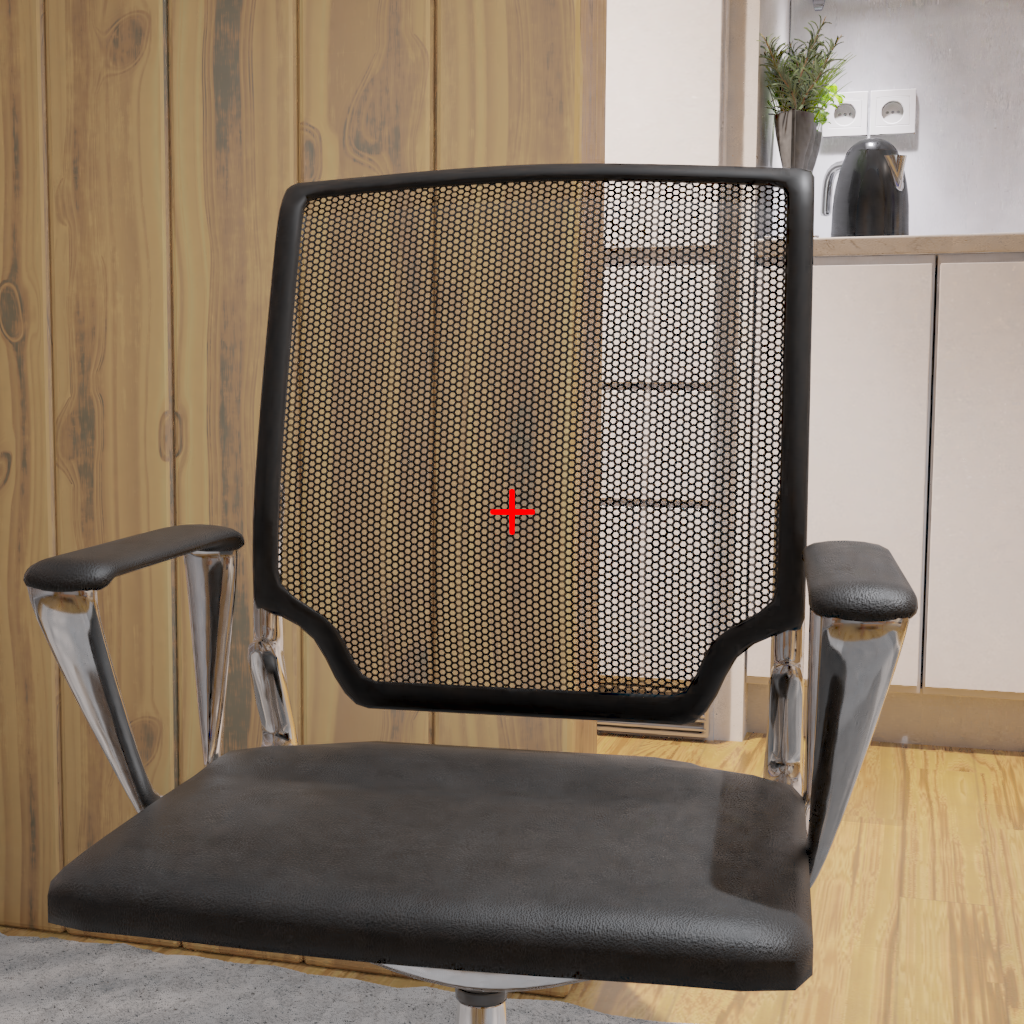}
        \includegraphics[width=0.495\textwidth]{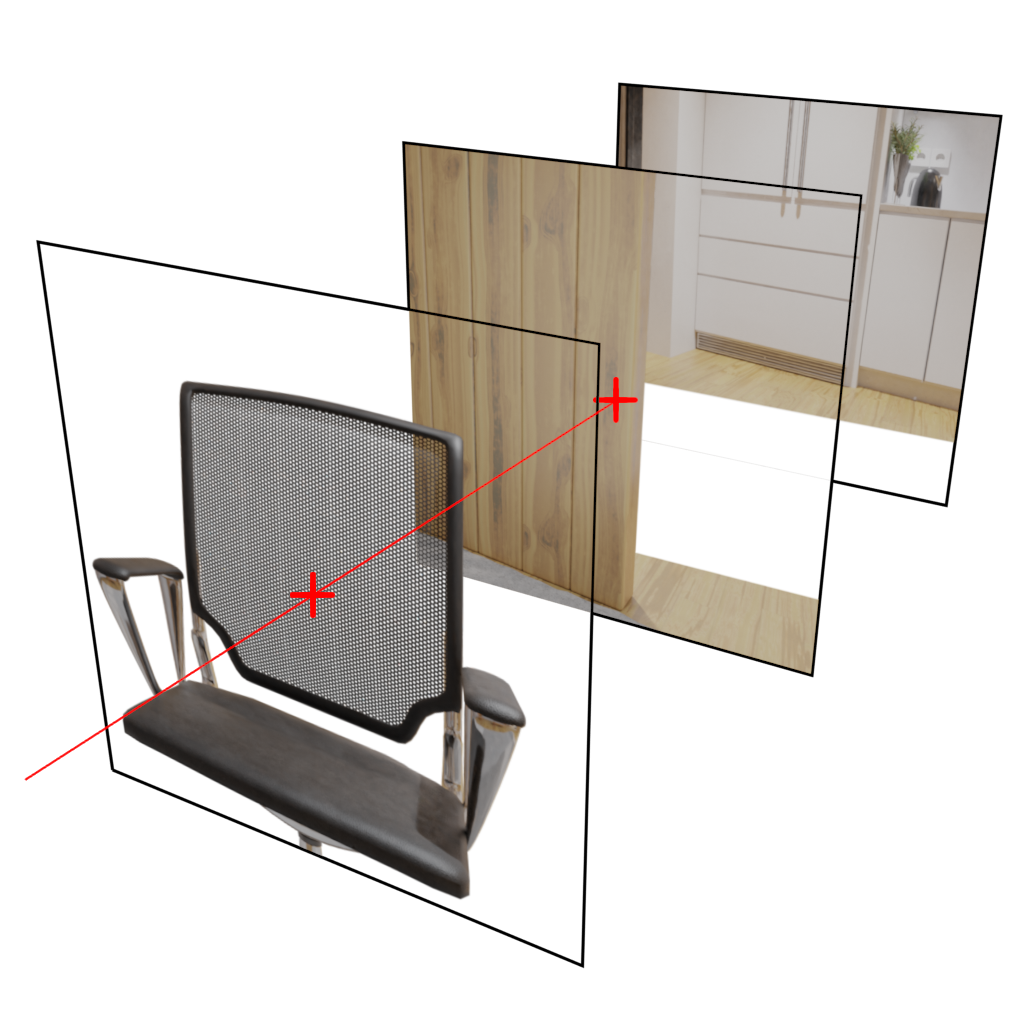}
        \subcaption{Rendered scene (left) containing multiple disparity layers (right)}
        \vspace{0.3cm}
    \end{subfigure}
    \begin{subfigure}{0.495\textwidth}
        \resizebox{\linewidth}{!}{\input{figures/teaser/teaser.tex}}
        \subcaption{\label{fig:teaser_posteriors}Disparity posterior distributions, predicted by different methods}
    \end{subfigure}
    \caption{
        \textbf{Comparison of disparity posterior distributions}. 
        Synthetic scene (a) containing overlapping objects at different depths.
        (b) shows disparity posterior distributions, estimated by different methods, for a single pixel (red crosses in (a)).
        This pixel captured two disparity modes (mesh material of chair (foreground) and wooden wall (background)).
        Note, that the Unimodal Posterior Regression (UPR) network, which outputs the mean and width of a Laplacian distribution, makes a wrong and uncertain prediction.
        The EPI-Shift Ensamble (ESE) detects both valid modes near the ground truth (GT).
    }
    \label{fig:teaser}
    \vspace{-3mm}
\end{figure}
Light field recordings and their applications, like real time rendering for virtual reality or highly accurate depth estimation, have improved vastly in recent years.
However, while light field rendering methods handle transparent and reflective objects well, current depth estimation methods still perform poorly in those areas.
State-of-the-art depth estimation methods mainly fail in three corner cases:
at objects edges, semi-transparent and reflective surfaces.
All three are caused by multiple objects at different depths contributing to the projected color of a single pixel on the camera sensor.
Most existing models fundamentally ignore these cases and assume only one ``true'' depth for each pixel.


Instead, we propose a series of deep-learning based methods to perform multimodal depth estimation, and depth estimation with uncertainty estimates.
For this, we start with the basic idea of outputting Bayesian posteriors, whereas standard regression models just produce a single estimate. 
From this idea, and by using a simple and well-founded maximum likelihood training framework, we develop three different light field depth estimation methods, all three of which are able to infer uncertainty estimates, and use multiple ground truth values during training. 
Two of the proposed methods are also able to predict multiple distinct depth values per pixel at test time.

To train our methods, we propose to utilize a multimodal dataset that contains the exact depth, color and opacity of all depth planes that are visible in an image. 
This is, in contrast to other current datasets which only contain a single ``true'' depth value.
Our multimodal dataset consists of randomly generated synthetic scenes with a significant proportion of occlusion and transparent objects.
This, for the first time, enables supervised training of multimodal depth estimation.

\hfill \newline
Our main contributions are as follows:
\begin{itemize}
\item 
An exploration of our three novel deep learning methods for light field based depth estimation being able to handle multiple depth modes:
\emph{(i)} Unimodal Posterior Regression (UPR); \emph{(ii)} EPI-Shift-Ensemble (ESE); \emph{(iii)} Discrete Posterior Prediction (DPP)
\item The release of the first multimodal light field dataset, containing the depth of all objects and their contribution to the color of each pixel in an image
\item A thorough evaluation of the predicted depth posterior distributions. We observe that the more restrictive UPR method works well when the unimodal depth assumption of traditional methods holds. In cases where it does not, the model is able to express a high uncertainty. In general, the discrete DPP method is superior to UPR and ESE.
\end{itemize}

%% file: figures/teaser/teaser.tex
\begin{tikzpicture}
    \begin{axis}[
        width=\linewidth,
        height=0.5\linewidth,
        legend cell align={left},
        grid=major,
        grid style={densely dotted, gray!50},
        legend style={fill=white, fill opacity=0.6, draw opacity=1,text opacity=1},
        x label style={at={(axis description cs:0.5,0.1)}},
        y label style={at={(axis description cs:0.125,0.5)}},
        ylabel=$p(y | x)$,
        xlabel=disparity $y$,
        xmin=-2,
        xmax=0.25,
        ymin=-0.1,
        ymax=3.6,
        yticklabels={,,},
        xticklabels={,,},
        ]
        
        \addplot[mark=none, my-orange, thick] 
        table[x=y,y=p,col sep=comma] {figures/teaser/upr.csv};
        
        \addplot[mark=none, my-green, thick] 
        table[x=y,y=p,col sep=comma] {figures/teaser/ib-inn.csv}; 

        \addplot[mark=none, my-blue, thick] 
        table[x=y,y=p,col sep=comma] {figures/teaser/ground_truth.csv}; 
        
        \addlegendentry{\scriptsize{UPR}}
        \addlegendentry{\scriptsize{ESE}}
        \addlegendentry{\scriptsize{GT}}
    \end{axis}
\end{tikzpicture}

%% file: sections/related_work.tex
In this section, we first discuss the related methods for depth estimation from light fields.
We secondly introduce different posterior prediction methods mainly inspired by works on uncertainty quantification.

\subsection{Light Field and Multimodal Depth Estimation}
Sinha \etal~\cite{sinha2012image} separate albedo and reflections in multi-view stereo recordings.
Their method discretizes the disparity space and utilizes a plane-sweep framework to compute an error volume based on pairwise normalized cross correlation.
Up to two distinct layers are extracted from this volume using a modified semi global-matching algorithm.

Johannsen \etal~\cite{johannsen2016sparse} introduced a similar method for multimodal light field depth estimation.
The method uses a dictionary of small EPI patches to encode light fields and estimate the depth at each pixel.
In addition, a mask that separates unimodal from bimodal pixels is computed.
Both aforementioned methods are able to estimate multiple depth modes from light fields but can not compete with the overall performance of state-of-the-art deep-learning based frameworks.
We refer to~\cref{sec:additional_experiments} for a comparison.

Heber \etal~\cite{heber2016convolutional} introduced the first deep learning based approach for light field depth estimation.
A neural network predicts the slope of local 2D per-pixel hyper planes.
In a second optimization step, a disparity map is inferred from the hyperplane parameters.
Due to noise and inaccurate results in untextured areas, an additional post-processing optimization is required.
The approach is a first step in the direction of learned depth from lightfields, but the results suffer from artifacts and blur which are addressed in their follow-up work~\cite{heber2016u,heber2017neural}.
Shin \etal~\cite{shin2018epinet} use a fully-convolutional architecture for direct disparity regression, consisting of two parts:
A multi-stream network comprised of four input networks for the horizontal, vertical and both diagonal light field view stacks.
The output features of those streams are concatenated and refined by a fully-convolutional head that directly outputs disparities.
EPI-Net achieves state-of-the-art performance on the HCI 4D Light Field Benchmark~\cite{honauer2016dataset} but is limited to a small disparity interval due to its small receptive field.

Leistner~\etal \cite{leistner2019learning} addressed the problem of light field depth estimation for high resolution and wide baseline light fields utilizing neural networks.
Instead of increasing the receptive field of the network which would yield worse generalization, they proposed to transform the input images with a shear transformation, called EPI-Shift.
The method applies a certain number of forward passes with shifted input EPIs and joins the result into a single prediction.

The most recent works by Tsai \etal~\cite{tsai2020attention} and Chen \etal~\cite{chen2021attention} aim for a better utilization of the light field data.
In contrast to previous methods, they utilize all light field views but fuse them in an early stage to reduce redundancy.

With Neural Radiance Fields (NeRF) Mildenhall \etal~\cite{mildenhall2020nerf} started a new line of research that represents the plenoptic function of a given scene as a neural network.
The network is trained using images recorded from different view points.
At first glance, the radiance along a ray in the NeRF framework also models a depth posterior distribution.
However, we argue that this assumption does not hold in general.
Zhang \etal~\cite{zhang2020nerf++} observed an inherent shape-radiance ambiguity in NeRF models.
This is a caused by training with a photometric loss only instead of supervised learning using ground truth depth.
As a consequence, there exists a family of radiance fields which perfectly explain all training images, even if the shape is incorrect.

Tosi \etal~\cite{tosi2021smd} addressed the problem of smoothness bias at depth discontinuities for stereo depth estimation methods. Their method predicts a bimodal Laplacian mixture distribution for each pixel and always chooses the mode with the largest probability density to retain sharp edges.

\subsection{Prediction of Regression Posteriors}
Traditional depth estimation methods predict a single depth value per pixel.
In the following, we therefore highlight works, focused on other tasks, that estimate a whole posterior instead.
MacKay~\cite{mackay1992practical} introduced the idea of Bayesian Neural Networks (BNNs) by proposing a probabilistic interpretation of the training process.
Instead of predicting only one output, the proposed network models the likelihood of possible outputs given the input.
The posterior probability of network weights, given the dataset can be inferred by maximizing this likelihood.
Now, Bayes rule is applied using maximum likelihood and some prior distribution over the network weights.
However, for large networks, this cannot be computed efficiently.

Neal~\cite{neal2012bayesian} introduced the Markov Chain Monte Carlo (MCMC) method to approximate the posterior over the model weights.
In addition, he analyzed the importance of the chosen prior over a high number of model weights.
This methods enabled an efficient approximation and therefore the training of BNNs.
However, even recent MCMC approaches are only applicable to a limited number of dimensions.


Kendall and Gal~\cite{kendall2017uncertainties} analyzed different types of uncertainty relevant for Computer Vision.
They propose to capture aleatoric uncertainty (uncertainty inherent in the data) by training the network to predict a variance for its output.
Epistemic uncertainty (uncertainty inherent in the model), which can be explained away using an infinite amount of data, is inferred using the Monte-Carlo Dropout technique~\cite{gal2016dropout}.
The authors propose a single model to infer both types of uncertainties. 

Ilg~\etal~\cite{ilg2018uncertainty} compared different uncertainty quantification methods for optical flow.
They compare aleatoric uncertainty learned by a single network to an ensemble of such networks and further, to a single network with multiple heads trained by the Winner-Takes-it-All loss.
In comparison, the authors find that Monte-Carlo Dropout does not work well for regression tasks and that ensembling also does not provide significantly better uncertainties than single networks.
However, we argue that this is caused by the lack of epistemic uncertainty for optical flow, due to the utilized synthetic datasets being quite large.
As shown by~\cite{kendall2017uncertainties}, Monte Carlo dropout ensembles capture epistemic uncertainty, which can be explained away given enough data.
Because epistemic uncertainty should be close to zero for huge synthetic datasets as used by~\cite{ilg2018uncertainty}, there is no room for improvement over a single network in the first place.

%

%% file: sections/method.tex
\newcommand{\shift}{\mathit{shift}}
Before we introduce our posterior estimation methods, we explain the basics of light field depth estimation.
A 4D light field is typically recorded by a 2D array of cameras, aligned on a regular grid.
Each camera is assigned a pair of so-called view coordinates $(u, v)$, \eg $(1, 1)$ for the top left camera.
Each pixel in an image is then also assigned a pair of image coordinates $(s, t)$.
A slice along one axis in image space with the corresponding axis in view space, \eg fixing $u$ and $s$, forms a so-called Epipolar Plane Image (EPI).
Each pixel in the central camera view is visible as a line structure in the EPI.
The slope of this line is equal to the negative inverse disparity of the pixel.
Thus, the task of depth estimation from light fields is to robustly detect this slope.
However, this approach assumes opaque, smooth and lambertian surfaces.
Non-textured, specular or semi-transparent regions and depth edges are ambiguous and therefore challenging even for state-of-the-art methods.
In this work we make progress for these cases by estimating the full depth posterior distribution.
This is especially useful for pixels with more than one valid depth mode, caused by either semi-transparency or the point spread at depth edges.
Unlike previous works that only predict a single depth, we are able to find those modes.



\subsection{Posterior Estimation}\label{sec:uncertainty}
From here on, we will simply refer to the input of a depth estimation network as $x$,
and to disparity as $y$.
In practice, the input to a depth estimation network is a concatenation of horizontal, vertical and diagonal EPIs.
Standard regression models usually output a single guess for the disparity, $\hat y = f_w(x)$,
where $f_w$ could be EPI-Net~\cite{shin2018epinet} with network weights denoted as $w$.
Instead, our goal is to estimate the posterior distribution $p(y\mid x)$ of the disparity $y$ given an input light field $x$.
In the following, we present four different approaches that all model such a posterior distribution.

To supervise more complex posterior distributions that can represent more than one mode, we created our own multimodal depth dataset.
Unlike common datasets that only contain a single ground truth disparity $y_i$ for each pixel $i$, we include the
disparities of multiple depth modes $y_{ij}$ for transparent objects and depth edges.
For each disparity we also include the amount of color $\eta_{ij}$ that it contributed to the pixel, 
\ie the perceived opacity of that object in the pixel.
From a Bayesian perspective, we interpret $\eta_{ij}$ as the probability $p(y_{ij})$ of this disparity.
This choice is justified in both an intuitive and methodological sense:
$\eta_{ij}$ corresponds to the fraction of the pixel's area that is taken up by the object at depth $y_{ij}$.
Lacking any prior knowledge, this is also equal to the probability that the depth at any subpixel position corresponds to that object.
This equality between the opacity $\eta_{ij}$ and probability $p(y_{ij})$ 
is valid both at edges as well as fine structures such as grids or woven meshes.
We also extend the definition to apply to semi-transparent materials such as printed glass as a simplifying assumption.
In Appendix \ref{sec:opacity_probability}, we give a more rigorous examination of the connection between $\eta_{ij}$ and $p(y_{ij})$.


\subsubsection{Unimodal Posterior Regression}
The most common approach for learning distributions is Maximum-Likelihood (ML) learning,
which most loss functions can be reformulated as.
The ML objective aims to find the model parameters $w$ which maximize the log likelihood of the training data 
$\{ (x_i, y_i) \}_{i=0}^N$ under the estimated posterior distribution.
In practice, we {minimize} the \emph{negative} log likelihood instead:
\begin{equation}
  \mathcal{L}_{\mathit{\mathrm{ML}}}  = - \frac{1}{N} \sum\limits_{i} \log p( y_i \mid x_i, w).
\end{equation}
It can be shown that this objective minimizes the Kullback-Leibler divergence (KLD) between $p( y \mid x, w)$
and the true posterior $p^*(y \mid x)$.

Previous regression based approaches \cite{shin2018epinet} simply use the $\mathcal{L}_1$ loss to make a single prediction:
\begin{equation}
  \mathcal{L}_{1}  = \frac{1}{N} \sum\limits_{i} | y_i - f_w(x_i) | .
\end{equation}
We see that this is equal to the ML objective when the posterior is assumed to be a Laplace distribution 
$p(y \mid x, w) \propto \exp( - | y - \mu | / b) / 2b$ with the network output $\mu = f_w(x)$ and a fixed value of $b=1$.
This motivates the following simple extension, 
which is an adaptation of the Dawid-Sebastiani-score \cite{dawid1999coherent}, later popularized in \cite{kendall2017uncertainties},
except that the $\mathcal{L}_2$-loss corresponding to a Gaussian posterior was used instead:
we allow the network to change the width $b$ of the posterior. With this, it becomes 
\begin{equation}
p(y \mid x, w) = \frac{1}{2b} \exp\left( - \frac{| y - \mu|}{b}\right), \; \text{with} \quad [b,\mu] = f_w(x).
\end{equation}
Putting this back into the ML objective, we get the following loss function for the predictive uncertainty:
\begin{equation}
    \mathcal{L}_{\mathrm{UPR}} = \frac{1}{N} \sum\limits_{i} \frac{|\mu_i-{y}_i|}{b_i} + \log b_i,
    \; \text{with} \quad [\mu_i, b_i] = f_w(x_i).
    \label{eq:loss_upr}
\end{equation}

This loss can be understood intuitively:
If the network struggles to predict $y_i$, the $\mathcal{L}_1$ loss term can 
be down-weighted by increasing the scale parameter $b$ for this pixel.
To avoid the trivial solution $b_i\to \infty$ for any input, high $b$ are penalized by a regularization term $\log b$.
In practice, we let the network predict $\log b$ instead of $b$ to improve numerical stability.

This approach gives us a measure of aleatoric uncertainty~\cite{kendall2017uncertainties} for each pixel which is already helpful for many downstream applications.
However, the implicit assumption for the method to work well is that the true posterior is also Laplacian.
Needless to say, this is certainly not true in multi-modal cases, which cannot be modeled by the Laplace distribution.
With multiple ground-truth depth modes $y_{ij}$ as opposed to only $y_i$, as in our dataset,
the loss for a Laplace distribution becomes 
\begin{equation}
    \mathcal{L}_{\mathrm{UPR}}^{\mathrm{MM}} = \frac{1}{N} \sum\limits_{i} \sum\limits_{j} p(y_{ij}) \frac{| y_{ij} - f_w(x_i) |}{\log b_i} + \log b_i
\end{equation}
and can be applied to the $\mathcal{L}_1$ loss respectively:
\begin{equation}
    \mathcal{L}_{1}^{\mathrm{MM}} = \frac{1}{N} \sum\limits_{i} \sum\limits_{j} p(y_{ij}) | y_{ij} - f_w(x_i) |
\end{equation}
However, in any case, those networks will focus on a single mode, or lie in between, and compensate for its wrong prediction
by expressing a very high uncertainty like in~\cref{fig:teaser}.

\subsubsection{EPI-Shift-Ensemble}
Commonly, one way to circumvent this exact issue is to use an ensemble of networks.
Instead of just estimating a single posterior, $M$ networks predict $M$ different posteriors, which are then averaged:
\begin{equation}
    p(y | x ) = \frac{1}{M}\sum\limits_k p(y | x, w_k)
    \label{eq:monte_carlo}
\end{equation}
over all networks with learned weights $w_k$, $k=1\dots M$.
It has been shown that ensembles deliver some of the best uncertainty estimates among existing methods \cite{ovadia2019can}.
Their main limitation is the high computational cost, especially for training.
Various approaches try to avoid this and train only a single model. 
For instance, Monte Carlo dropout exhibits similar characteristics as a true ensemble \cite{hara2016analysis}.
Instead, we propose a new scheme which uniquely exploits the nature of light field data, which we term EPI-Shift-Ensemble (ESE).
To motivate this, we take note of the technique from \cite{leistner2019learning}, 
where the EPI is sheared in such a way that a global offset $\Delta y$ is added to the disparity of the EPI.
This is advantageous, because it enables inference on large disparities \eg for wide-baseline light field cameras.
We denote this operation as $\shift(x, \Delta y)$ and extend it to arbitrary sub-pixel steps. 
To form our EPI-Shift-Ensemble, we successively shift the input $M$ times in steps of $\Delta y$.
This gives us $M$ different augmented inputs, where each has a different artificial disparity offset.
Each input is fed into the network, and the resulting posteriors are shifted back by the same offset.
In this way, we get $M$ different estimates by the same network, each from a different input containing the same information.
The overall posterior can be expressed as
\begin{equation}
    p_\mathit{ESE}(y \mid x, w ) = \frac{1}{M}\sum\limits_{k}
    p\Big(y - k\cdot \Delta y \Big| \shift(x, k \cdot \Delta y) , w\Big).
\end{equation}
In the summation, $k= -\lfloor M/2 \rfloor \dots \lceil M/2 \rceil$. 
For the shape of the individual posteriors, we use the same Laplacian as before.
Note that the network weights $w$ are shared for each summand, only the shift is different.
However, this operation alone does not prevent the problems seen for the single mode approach:
the network will try to average out bi-modal solutions, or collapse into one mode.
As a result, we see little to no diversity in the EPI-Shift-Ensemble, and no multi-modal posteriors.
To prevent the collapse, we mask the loss during training so that it only applies to pixels with a shifted $y' = y - k \cdot \Delta y$ within the range of one step $|y'| < \Delta y / 2$.
In all other cases the output will have a large uncertainty:
\begin{equation}
\mathcal{L}_{\mathrm{ESE}} = 
    \frac{1}{N}\sum\limits_{i}
    \begin{cases} 
        \frac{|\mu_i-{y}_i|}{b_i} + \log b_i & \text{if } |{y_i'}| < \frac{\Delta y}{2} \\
        0 & \text{otherwise.}
    \end{cases}
    \label{eq:loss_ese}
\end{equation}
We extend this to our multimodal dataset similarly to our unimodal networks:
\begin{equation}
\mathcal{L}_{\mathrm{ESE}}^\mathrm{MM} = 
    \!\frac{1}{N}\! \sum\limits_{i}\sum\limits_{j}
    p(y_{ij})
    \begin{cases} 
        \frac{|\mu_{i}-{y}_{ij}|}{b_{i}} \!+\! \log b_{i} \!\! & \!\! \text{if } |{y_{ij}'}| < \frac{\Delta y}{2} \\
        0 & \text{otherwise.}
    \end{cases}
    \label{eq:loss_ese_mm}
\end{equation}
After training with this loss, the network will only be confident (narrow posterior) if 
it estimates that the input disparity is $y \in [-\Delta y / 2, \Delta y / 2]$,
and the predicted posterior will always be centered in this range.
When using this network in the EPI-Shift-Ensemble, 
we see that each term $k$ will only contribute a narrow posterior if a plausible disparity
lies between $(k - 1/2) \Delta y$ and $(k + 1/2) \Delta y$, 
thus ensuring diverse outputs and the possibility of multi-modal predictions.
Three details should be noted:
First, the model is not trained as an ensemble, a single model is trained just as before with the modified $\mathcal{L}_\mathrm{ESE}$-loss.
The ensembling operation is only performed at inference time.
Second, the masked loss does not reduce the effective size of our training set, as we also apply random EPI-shifts as a part of the data augmentation process.
This way, all pixels will randomly fulfill $|{y_i}| < \Delta y / 2$ at some point.
Lastly, the inference time is $M$ times longer, as $M$ forward passes have to be performed to compute the EPI-Shift-Ensemble.

\subsubsection{Discrete Posterior Prediction}
The approach of discretizing regression tasks has been successful for stereo depth estimation in the past,
and promises to model more expressive posteriors.
Specifically,
by discretizing the range of disparities, a softmax output can be used to represent the posterior.
The posterior is then a step function consisting of these discrete probabilities.
If $y_j$ are the discretization steps, we can write
\begin{equation}
    p(y_j|x) \propto \mathit{softmax}\left(f_w(x_i)\right)_j \coloneqq
    \frac{\exp\left( f_w(x_i)_j \right)}
    {\sum\limits_k \exp\left(f_w(x_i)_k\right)}
\end{equation}
If multiple modes are used for the training, we also discretize the distribution $p(y_j)$ over these modes.
Maximum likelihood training is then simply equivalent to the categorical cross-entropy (CE) loss,
where the correct ``class'' is the bin $j$ that the training example $y_i$ lands in:
\begin{equation}
    \mathcal{L}_{\mathrm{CE}} = \frac{1}{N} \sum_i
    - \log \Big(
    \mathit{softmax}\big(f_w(x_i)\big)_j \Big).
    \label{eq:loss_ce}
\end{equation}
For the multimodal dataset, we can compute the cross-entropy accordingly:
\begin{equation}
    \mathcal{L}_{\mathrm{CE}}^{\mathrm{MM}} = \frac{1}{N} \sum_i \sum_j
    - p(y_j) \log \Big(
    \mathit{softmax}\big(f_w(x_i)\big)_j \Big).
\end{equation}
Note, that the two correspond, for a unimodal dataset, we simply have $p(y_j) = 1$ and $p(y_{l \neq j}) = 0$,
simplifying to the cross-entropy in Eq. \ref{eq:loss_ce}.

Although the discrete posterior prediction gives us an uncertainty estimation at much lower computational cost than the EPI-Shift-Ensemble 
and can represent more flexible posteriors compared to simple Laplacians,
softmax probabilities are generally known to be overconfident \cite{guo2017calibration}
and also make wrong but confident predictions in ambiguous or unseen cases.
Different techniques for post-calibration of uncertainties exist, outlined in \cite{guo2017calibration}.
However, while they may prevent overconfidence going from the training to a test set, 
they do not make the uncertainties more reliable in general \cite{ovadia2019can},
\eg for ambiguous inputs.

\subsection{Dataset Generation}
\begin{figure}[t]
    \centering
    \begin{subfigure}{0.325\linewidth}
        \includegraphics[width=\textwidth, trim=0 0 0 80, clip]{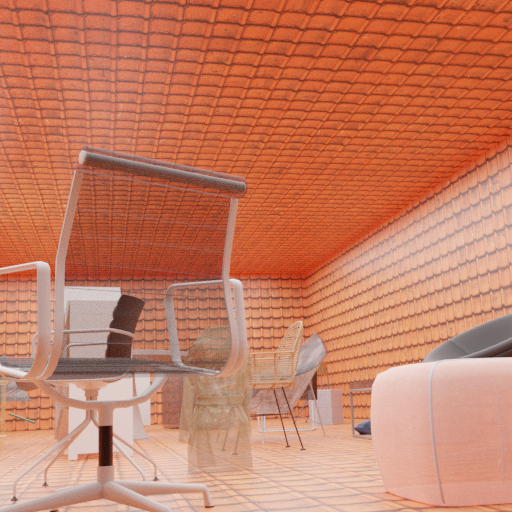}
        \subcaption{
            Exemplary scene
        }
    \end{subfigure}
    \begin{subfigure}{0.325\linewidth}
        \includegraphics[width=\textwidth, trim=0 0 0 80, clip]{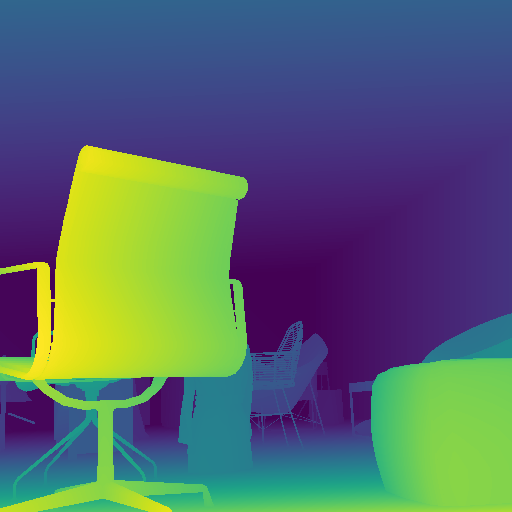}
        \subcaption{
            First disparity
        }
    \end{subfigure}
    \begin{subfigure}{0.325\linewidth}
        \includegraphics[width=\textwidth, trim=0 0 0 80, clip]{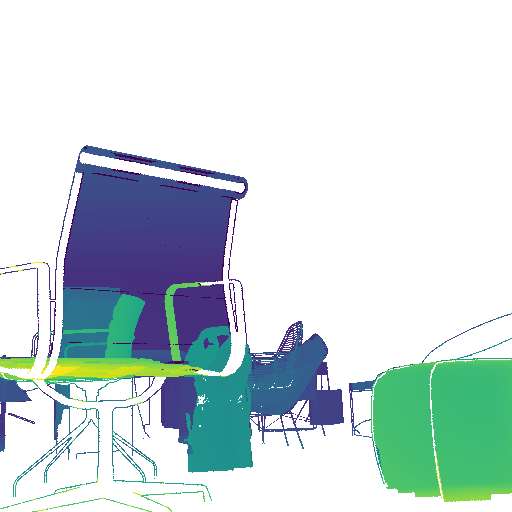}
        \subcaption{
            Second disparity
        }
    \end{subfigure}
    \caption{
        An exemplary scene (a) from our randomly generated dataset. (b) and (c)  show different disparity modes.
        Each pixel has at least one disparity mode.
        Behind semi-transparent objects and at depth edges, a second disparity mode (c) exists.
    }
    \label{fig:dataset}
    \vspace{-3mm}
\end{figure}
To train and validate all methods above, ground truth multimodal depth data is required.
Because all previous light field datasets only contain a single depth per pixel, we generated a novel multimodal depth light field dataset containing 110 randomly generated indoor scenes (see \cref{fig:dataset}).
To improve the training performance, the dataset generator follows four goals: \textit{(i)} relatively photorealistic appearance, \textit{(ii)} high diversity to improve generalization of trained models, \textit{(iii)} many occlusions and depth edges to improve the performance at object edges and, \textit{(iv)} a large proportion of pixels with multiple valid depths.
Note, that we decided to not include reflections explicitely.
However, as~\cite{sinha2012image} demonstrates, reflections at flat surfaces behave similar to alpha-transparency.
Therefore, the methods trained with our dataset should also be usable to estimate multimodal depths caused by reflections.
We refer to~\cref{sec:dataset_generation} for more information on the dataset generation process.

%% file: sections/experiments.tex
In this section, we describe our training procedure and analyze the predicted posterior distributions.
We therefore distinguish two possible applications: unimodal prediction with uncertainty and multimodal prediction.

\subsection{Architectures and Training Details}
\begin{figure}[t]
    \centering
    \begin{minipage}{.55\linewidth}
        \begin{subfigure}{\linewidth}
            \centering
            \resizebox{0.6\linewidth}{!}{\input{figures/architectures/base}}
            \subcaption{Baseline}
            \vspace{0.25cm}
        \end{subfigure}
        \begin{subfigure}{\linewidth}
            \centering
            \resizebox{0.6\linewidth}{!}{\input{figures/architectures/upr}}
            \subcaption{Unimodal Posterior Regression}
            \vspace{0.25cm}
        \end{subfigure}
        \begin{subfigure}{\linewidth}
            \resizebox{\linewidth}{!}{\input{figures/architectures/dpp}}
            \subcaption{Discrete Posterior Prediction}
        \end{subfigure}
    \end{minipage}
    \hfill
    \begin{minipage}{.4\linewidth}
        \begin{subfigure}{\linewidth}
            \vspace{-0.1cm}
            \resizebox{0.8\linewidth}{!}{\input{figures/architectures/ese}}
            \subcaption{EPI-Shift Ensemble}
        \end{subfigure}
    \end{minipage}
    \caption{
        \textbf{Network architecture overview:} 
        Our baseline is a simple feed-forward network trained to only predict disparity (a).
        We compare three posterior-regression methods:
        Laplacian prediction using learned loss attenuation (b), an ensemble on shifted inputs (d) and a discrete softmax classification network (c).
    }
    \label{fig:models}
    \vspace{-3mm}
\end{figure}
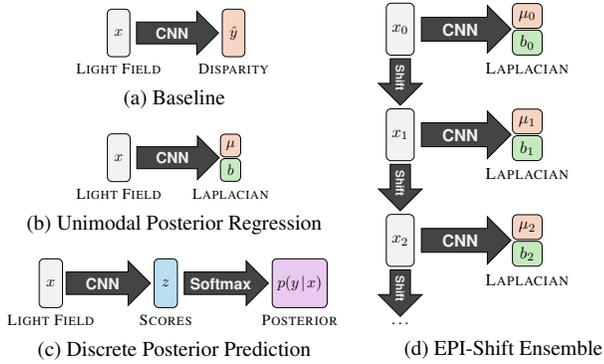
To ensure a fair comparison, we chose the state-of-the-art method EPI-Net~\cite{shin2018epinet} with minor modifications as our backbone network architecture for all models.
We describe the exact network architectures in more detail in~\cref{sec:architectures}.
The network consists of a total number of eight blocks, containing two convolutional layers, followed by a Rectified Linear Unit (ReLU) each and one Batch Normalization layer.
For UPR, we added an additional output layer to also predict the negative log width $-\log b$.
The EPI-Shift Ensemble is a modification of UPR with the only difference being the loss functions.
We experimented with different $\Delta y$ for our ESE method, resulting in different $M$s.
$\Delta y = 0.1$ performed best and gives $M = 70$ for the disparity range of our data.
Our extended EPI-Shift transformation, based on~\cite{leistner2019learning}, is described in more detail in~\cref{sec:epi-shift}.
For the discrete DPP method, we chose a number of $108$ ``classes''.
This is motivated by the common BadPix007 metric that consideres a pixel as ``correct'' if it is closer than $0.07\mathrm{px}$ to the ground truth.
A number of $108$ classes in a disparity range of $-3.5\dots3.5$ leads to a bin size of $\approx 0.065$ which is slightly below this threshold.
We implemented  our framework using PyTorch~\cite{paszke2019pytorch}.

We trained all networks using the loss functions described in \cref{sec:uncertainty}.
The unimodal loss functions, denoted as $\mathcal{L}_x$ are always applied to the closest disparity.
All multimodal loss functions, denoted as $\mathcal{L}_x^{\mathrm{MM}}$ are applied to all disparity modes.

In addition, we reimplemented EPI-Net~\cite{shin2018epinet} in our own framework as a baseline for a fair comparison.
The learning rate was set to $10^{-3}$ for all models, using a batch size of $512$ and the Adam optimizer.
We trained on randomly cropped patches ($96px \times 96px$) from the set of $100$ training scenes in our dataset.
To further improve the diversity of our dataset, we make use of a number of data augmentation operations:
We apply a random sub-pixel EPI-Shift between $-2\mathit{px}$ and $2\mathit{px}$~\cite{leistner2019learning}.
In addition, we randomly rotate the light field by a multiple of $90^\circ$, randomly rotate the colors in RGB-space and randomly change brightness and contrast.

\subsection{Posterior Evaluation}
Depth estimation methods are usually evaluated by measuring the pixel-wise error to a ground truth disparity map.
However, to correctly measure the quality of estimates in areas with multiple valid depths, a different set of metrics is required.

We consider two application scenarios:
\textit{(i)} the estimation of just a single disparity, but with an additional confidence measure to ensure that the estimate can be ``trusted''.
This may be required, \eg in industrial and robotics applications where decisions are based on the estimated depth;
\textit{(ii)} the estimation of multiple depths in areas with transparent objects or at object edges.
This is typically required by Computer Graphics applications that aim to render the recorded scene from a different angle.
In the following, we introduce metrics for both cases.


\subsubsection{Unimodal Prediction with Uncertainty}
\begin{figure*}[t]
    \centering
    \begin{subfigure}{0.495\textwidth}
        \resizebox{\linewidth}{!}{\input{figures/sparsify/cnn.tex}}
        \subcaption{Sparsification curve}
        \label{fig:sparsify_curve}
    \end{subfigure}
    \begin{subfigure}{0.495\textwidth}
        \resizebox{\linewidth}{!}{\input{figures/sparsify/error.tex}}
        \subcaption{Sparsification Error for all methods\label{fig:sparsify_error}}
    \end{subfigure}
    \caption{
        \textbf{Unimodal uncertainty quantification:} Sparsification results of analyzed methods with respect to the disparity BadPix007.
        By removing a certain fraction with the highest predicted uncertainty (a), the error decreases.
        The ``Oracle'' is a lower bound, created by removal of truly worst pixels.
        We also compute the difference between the predicted sparsification and its ``Oracle'', denoted as Sparsification Error (SE) (b), for a comparison of the three methods (trained on all depth modes)
    }
    \label{fig:sparsify}
    \vspace{-3mm}
\end{figure*}
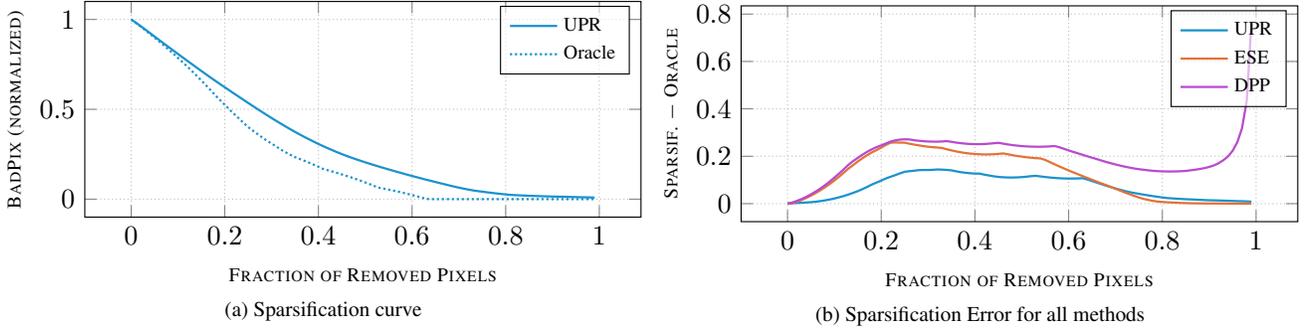
Previous methods and datasets~\cite{honauer2016dataset} always consider the disparity of the closest object as ``true'', even when this object is transparent.
However, estimation methods oftentimes output the disparity of the background object in those cases, which may lead to severe issues in downstream applications.
In addition, more ambiguities usually occur in non-textured areas which cannot be estimated correctly.
Most optimization-based methods fill in those ambiguous regions by interpolation between adjacent pixels with confident predictions.
Due to the limited receptive field of neural network based methods, this is only possible to some extent.
In any case, a confidence measure is extremely useful for downstream applications in order to decide if an estimate can be trusted.
To achieve this, the overall variance $\sigma^2$ of the predicted posterior distribution can be used as an uncertainty measure.
We aim for a consistently high uncertainty in regions with ambiguous predictions.
To evaluate the quality of the estimated posteriors,
we remove the $s\%$ of pixels with the highest posterior variance (uncertainty).
With these ambiguous and uncertain cases filtered out, the BadPix of the remaining pixels is lower,
which can be plotted as a sparsification curve (see \cref{fig:sparsify_curve}).
The optimal curve can be computed by removing those $s\%$ of pixels with the largest ground truth error.
We call this the ``oracle'' curve, 
which represents the lower bound of what is achievable.
This method is commonly used to evaluate uncertainties for regression tasks~\cite{ilg2018uncertainty}.
In order to compare all methods, we compute the Sparsification Error (SE) by subtracting the oracle curve from the sparsification curve.
The Area under the Sparsification Error (AuSE) quantifies the uncertainty quality of each method with a single number.

\subsubsection{Multimodal Prediction}
\newcommand{\columnSpace}{3pt}
\addtolength{\tabcolsep}{\columnSpace}
\begin{table*}[h]
    \begin{center}
        \begin{tabular}{l|r r | r r r | r | r}
        \hline
            \textbf{Method} &
            \multicolumn{2}{c|}{\textbf{Unimodal Metrics}} &
            \multicolumn{3}{c|}{\textbf{KL Divergence}} &
            \textbf{AuSE} $\downarrow$&
            \textbf{Time} $\downarrow$\\
            &
            MSE $\downarrow$&
            BadPix $\downarrow$&
            Unimodal $\downarrow$&
            Multimodal $\downarrow$&
            Overall $\downarrow$&
            &  (in sec)    \\
        \hline
        \hline
            BASE (uni) &
            \textbf{0.374} & 0.229 & 4.720 & 7.876 & 5.421 & - & \textbf{2.188} \\

            BASE (multi)  &
            0.563 & 0.307 & 5.259 & 8.514 & 6.025 & - & 2.211 \\
        \hline
            UPR (uni) &
            0.439 & 0.235 & 1.719 & 3.381 & 1.879 & \textbf{0.071} & 2.260 \\

            UPR (multi) &
            0.676 & 0.285 & 1.987 & 3.156 & 2.114 & 0.072 & 2.287 \\
        \hline
            ESE (uni) &
            1.269 & 0.223 & 4.164 & 3.628 & 4.160 & 0.099 & 17.492 \\

            ESE (multi) &
            1.850 & 0.229 & 4.283 & 3.719 & 4.277 & 0.121 & 16.902 \\
        \hline
            DPP (uni) &
            0.765 & \textbf{0.209} & \textbf{1.631} & 3.057 & \textbf{1.734} & 0.272 & 4.348 \\

            DPP (multi) &
            0.686 & 0.231 & 1.824 & \textbf{2.987} & 1.914 & 0.197 & 4.382 \\
        \hline
        \end{tabular}
    \end{center}
    \vspace{-3mm}
    \caption{
        \textbf{Evaluation}, from left to right:
        Mean Squared Error and the common BadPix007 score (percentage of pixels with $|y_i - \hat{y}_i| > 0.07$),
        Kullback-Leibler divergence on unimodal, multimodal and all pixels,
        Area under Sparsification Error (AuSE),
        runtime of one forward pass.
        Our methods were trained with both losses $\mathcal{L}_x$ (uni) and $\mathcal{L}_x^{\mathrm{MM}}$ (multi) respectively.
        Lower is better
    }
    \label{tab:results}
\end{table*}
For many applications, including rerendering of a recorded scene from different angles, estimating multiple depths for pixels at object edges and transparent surfaces is desirable.
In addition to the disparities of all modes, the contribution of each mode to the color of the pixel is also important.
To evaluate both, we measure the Kullback-Leibler divergence (KLD) 
\begin{equation}
    \mathcal{D}_{\mathrm{KL}} = \int p(y_i) \log \left( \frac{p(y_i)}{p(y_i | x_i)} \right)
\end{equation}
between the predicted disparity posterior $p(y_i | x_i)$ and the true disparity distribution $p(y_i)$ at a pixel $i$.
Intuitively, the KLD will be minimal if the posterior assigns a high probability density to each true disparity mode and a low density to disparities that are not present at a pixel $i$.
In other words: the KLD is lowest, if each mode is recovered sharply, instead of a broad uncertainty covering both.
E.g.\ in~\cref{fig:teaser_posteriors}, the green and orange curves have the same uncertainty, but the green has a much lower KLD.
Optimally, the density at each disparity mode corresponds to the contribution $\eta_i$ of this mode to the resulting pixel color.
As the KLD is only well defined between two continuous or two discrete distributions and we compare continuous as well as discrete methods, we chose to discretize all distributions.
We therefore assign each ground truth disparity to one out of $K = 108$ bins with a width of $h \approx 0.065\mathrm{px}$ each.
\begin{equation}
    p(y_k) =  \sum_j p(y_j) \; \forall j \; \mathrm{with} \; | y_j - y_k | < \frac{h}{2}
\end{equation}
This is, again, motivated by the well-established BadPix007 metric~\cite{honauer2016dataset} which considers a pixel as correct if the $\mathcal{L}_1$ distance to the ground truth is below $0.07\mathrm{px}$.
For the baseline method that only outputs one disparity, we simply set the probability of the bin that contains this disparity to $1$.
All continuous posterior distributions are discretized by integrating over the interval of each bin:
\begin{equation}
    p(y_k | x) = \int\limits_{(k-0.5) h}^{(k+0.5) h} p(y | x) \mathrm{d}y
\end{equation}
We now average the discrete KLD
\begin{equation}
    \mathcal{D}_{\mathrm{KL}} = \frac{1}{NK} \sum\limits_{i} \sum\limits_{k} p(y_{ik}) \log \left( \frac{p(y_{ik})}{p(y_{ik} | x_i)} \right)
\end{equation}
over all pixels in all validation scenes.
In addition, we also compute the KLD over all unimodal and all multimodal pixels separately. 
A pixel is considered multimodal if it has at least two modes $j$ with $p(y_j) > 0.3$.


\subsubsection{Results}
\Cref{tab:results} compares the unimodal, multimodal and sparsification performance of all methods.
In the following, we will interpret our results with respect to the aforementioned applications: unimodal disparity estimation with uncertainty and multimodal disparity estimation.

With respect to pure unimodal performance, our baseline method and DPP perform best.
The higher MSE for DPP is caused by small discretization errors due to the discrete number of bins.
Those small errors are well below a threshold of $0.07\mathrm{px}$ and therefore ignored by the BadPix metric
which shows that DPP indeed predicts $2\%$ more pixels correctly compared to the baseline.
UPR performs only slightly worse than both methods overall.
However, due to the uncertainty being directly supervised by $\mathcal{L}_{\mathrm{UPR}}$ this method is superior in terms of sparsification.
This means that its uncertainty metric reflects most accurately whether a prediction is correct.
We conclude that, if the application requires only a single disparity and confidence is important, UPR should be considered.

With respect to the accuracy of the predicted posterior distribution, DPP performs best in unimodal and also multimodal areas.
However, as most softmax prediction methods, it is overconfident, as reflected by the sparsification error in~\cref{fig:sparsify_error}.
Despite the popularity of ensemble-based models for uncertainty estimation, ESE cannot compete with the other two methods.
It especially performs worse in highly non-textured areas with a large and noisy uncertainty output $b_i$ for all ``ensemble members''.
To decide for a unimodal disparity estimate, the member with the lowest uncertainty is still used, but due to the noise, this choice becomes arbitrary.
The chosen member can only output a disparity from a small interval within the potentially large disparity range. 
On average, this causes a high deviation from the ground truth for ESE.
In contrast, our other methods are able to ``smooth'' the disparity in uncertain areas from adjacent, more certain pixels.
This is also reflected by the relatively high MSE but low BadPix error:
the amount of ``correctly'' predicted pixels is on par with other methods, but the deviation of ``wrongly'' predicted pixels is generally higher.
In addition, as all members contribute slightly to the mixture of Laplacians, the density of the posterior is higher along the whole disparity interval, which leads to a worse multimodal KLD compared to UPR and DPP.
We therefore generally recommend DPP for multimodal predictions in small, narrow-baseline light fields.
However, due to its shift-operation, ESE can, unlike other methods, operate on arbitrary large disparity ranges and is therefore still advisable for high-resolution or wide-baseline light field cameras.
In addition, it performs also relatively well in terms of sparsification.
Comparing the methods trained on only one mode with the same methods trained on multiple modes shows that multimodal training leads to a slightly better multimodal performance for UPR and DPP, but always comes at a cost in unimodal areas.
Our baseline method cannot efficiently represent multimodal posteriors as it only predicts a single disparity.

We conclude that the exact model and training method should be carefully chosen based on the intended application.
We refer to~\cref{sec:additional_experiments} for examples of estimated disparity posterior distributions, an evaluation of our methods on the common HCI 4D Light Field Dataset~\cite{honauer2016dataset}, additional qualitative results as well as a comparison to Sinha \etal~\cite{sinha2012image} and Johannsen \etal~\cite{johannsen2016sparse}.

%% file: figures/architectures/base.tex
\begin{tikzpicture}[
    data domain/.style={fill=black!20, draw=black, rounded corners=3pt, minimum height=3.15em, minimum width=1.5em, align=center, font=\normalsize},
    half data domain/.style={fill=black!20, draw=black, rounded corners=3pt, minimum height=1.5em, minimum width=1.33em, align=center, font=\normalsize},
    fat arrow/.style={double arrow, fill=my-gray, draw=black, align=center, text depth=-0.5pt, text width=3em, text=black!5},
    unidirectional fat arrow/.style={single arrow, fill=my-gray, draw=black, align=center, text depth=-0.5pt, text width=4em, text=black!5},
    loss/.style={black!75, midway, align=center, font=\footnotesize\sffamily\bfseries}]

    \node [data domain, fill=black!5] (x) {$x$};
    \node [below=6pt of x.south, yshift=5pt] (inp) {\normalsize{\textsc{Light Field}}};

    \node [unidirectional fat arrow, right=4pt of x.east] (cnn) {\textsf{\normalsize{\textbf{CNN}}}};
    \node [data domain, fill=black!5, right=2pt of cnn.east, fill=my-orange!30] (y) {$\hat y$};
    \node [below=6pt of y.south, yshift=5pt] (out) {\normalsize{\textsc{Disparity}}};
\end{tikzpicture}

%% file: figures/architectures/upr.tex
\begin{tikzpicture}[
    data domain/.style={fill=black!20, draw=black, rounded corners=3pt, minimum height=3.15em, minimum width=1.5em, align=center, font=\normalsize},
    half data domain/.style={fill=black!20, draw=black, rounded corners=3pt, minimum height=1.5em, minimum width=1.33em, align=center, font=\normalsize},
    fat arrow/.style={double arrow, fill=my-gray, draw=black, align=center, text depth=-0.5pt, text width=3em, text=black!5},
    unidirectional fat arrow/.style={single arrow, fill=my-gray, draw=black, align=center, text depth=-0.5pt, text width=4em, text=black!5},
    loss/.style={black!75, midway, align=center, font=\footnotesize\sffamily\bfseries}]

    \node [data domain, fill=black!5] (x) {$x$};
    \node [below=6pt of x.south, yshift=5pt] (inp) {\normalsize{\textsc{Light Field}}};

    \node [unidirectional fat arrow, right=4pt of x.east] (cnn) {\textsf{\normalsize{\textbf{CNN}}}};
    \node [half data domain, above right=0.5pt of cnn.east, fill=my-orange!30] (mu) {$\mu$};
    \node [half data domain, below right=0.5pt of cnn.east, fill=my-green!30] (sigma) {$b$};
    \node [below=6pt of sigma.south, yshift=5pt] (out) {\normalsize{\textsc{Laplacian}}};
\end{tikzpicture}

%% file: figures/architectures/dpp.tex
\begin{tikzpicture}[
    data domain/.style={fill=black!20, draw=black, rounded corners=3pt, minimum height=3.15em, minimum width=1.5em, align=center, font=\normalsize},
    fat arrow/.style={double arrow, fill=my-gray, draw=black, align=center, text depth=-0.5pt, text width=3em, text=black!5},
    unidirectional fat arrow/.style={single arrow, fill=my-gray, draw=black, align=center, text depth=-0.5pt, text width=4em, text=black!5},
    loss/.style={black!75, midway, align=center, font=\normalsize\sffamily\bfseries}]

    \node [data domain, fill=black!5] (x) {$x$};
    \node [below=6pt of x.south, yshift=5pt] (inp) {\normalsize{\textsc{Light Field}}};

    \node [unidirectional fat arrow, right=2pt of x.east] (INN) {\textsf{\normalsize{\textbf{CNN}}}};

    \node [data domain, right=2pt of INN.east, fill=my-blue!30] (z) {$z$};
    \node [below=6pt of z.south, yshift=5pt] (gmm) {\normalsize{\textsc{Scores}}};
    
    \node [unidirectional fat arrow, right=4pt of z.east] (inference) {\textsf{\normalsize{\textbf{Softmax}}}};
    
    \node [data domain, right=2pt of inference.east, fill=my-purple!30] (y) {$p(y \! \mid \! x)$};
    \node [below=6pt of y.south, yshift=5pt] (output) {\normalsize{\textsc{Posterior}}};
\end{tikzpicture}

%% file: figures/architectures/ese.tex
\begin{tikzpicture}[
    data domain/.style={fill=black!20, draw=black, rounded corners=3pt, minimum height=3.15em, minimum width=1.5em, align=center, font=\normalsize},
    half data domain/.style={fill=black!20, draw=black, rounded corners=3pt, minimum height=1.5em, minimum width=1.8em, align=center, font=\normalsize},
    fat arrow/.style={double arrow, fill=my-gray, draw=black, align=center, text depth=-0.5pt, text width=3em, text=black!5},
    unidirectional fat arrow/.style={single arrow, fill=my-gray, draw=black, align=center, text depth=-0.5pt, text width=4em, text=black!5},
    unidirectional fat short arrow/.style={single arrow, fill=my-gray, draw=black, align=center, text depth=-0.5pt, text width=1.5em, text=black!5},
    loss/.style={black!75, midway, align=center, font=\normalsize\sffamily\bfseries}]
    
    \foreach \i/\y in {0/0,1/-2.25,2/-4.5}
    {
        \node [data domain, fill=black!5] (x) at (0, \y) {$x_{\i}$};

        \node [unidirectional fat arrow, right=4pt of x.east] (cnn) {\textsf{\normalsize{\textbf{CNN}}}};

        \node [half data domain, above right=0.5pt of cnn.east, fill=my-orange!30] (mu) {$\mu_{\i}$};
        \node [half data domain, below right=0.5pt of cnn.east, fill=my-green!30] (sigma) {$b_{\i}$};
        
        \node [unidirectional fat short arrow, below=9pt of x.south, rotate=-90, xshift=4pt, yshift=6.375pt] (shift) {\textsf{\scriptsize{\textbf{Shift}}}};
        
        \node [below=6pt of sigma.south, yshift=5pt] (out) {\normalsize{\textsc{Laplacian}}};
    }
    
    \node [below=25pt of x.south, yshift=-5pt, xshift=0.75pt] (inp) {\dots};

\end{tikzpicture}

%% file: figures/sparsify/cnn.tex
\begin{tikzpicture}
    \begin{axis}[
        width=\linewidth,
        height=0.5\linewidth,
        legend cell align={left},
        legend style={fill=white, fill opacity=0.6, draw opacity=1,text opacity=1},
        grid=major,
        grid style={densely dotted, gray!50},
        x label style={at={(axis description cs:0.5,0.0)}},
        y label style={at={(axis description cs:0.05,0.5)}},
        ylabel=\scriptsize{\textsc{BadPix (normalized)}},
        xlabel=\scriptsize{\textsc{Fraction of Removed Pixels}},
        ]

        \addplot[mark=none, my-blue, thick] 
        table[x=frac,y=uncert,col sep=comma] {figures/sparsify/upr.csv}; 

        \addplot[mark=none, my-blue, thick, densely dotted]
        table[x=frac,y=oracle,col sep=comma] {figures/sparsify/upr.csv}; 
        

        \addlegendentry{\scriptsize{UPR}}
        \addlegendentry{\scriptsize{Oracle}}
    \end{axis}
\end{tikzpicture}

%% file: figures/sparsify/error.tex
\begin{tikzpicture}
    \begin{axis}[
        width=\linewidth,
        height=0.5\linewidth,
        legend cell align={left},
        legend style={fill=white, fill opacity=0.6, draw opacity=1,text opacity=1},
        grid=major,
        grid style={densely dotted, gray!50},
        x label style={at={(axis description cs:0.5,0.0)}},
        y label style={at={(axis description cs:0.05,0.5)}},
        ylabel=\scriptsize{\textsc{Sparsif. $-$ Oracle}},
        xlabel=\scriptsize{\textsc{Fraction of Removed Pixels}},
        ]

        \addplot[mark=none, my-blue, thick] 
        table[x=frac,y=sparse_err,col sep=comma] {figures/sparsify/upr.csv}; 
        
        \addplot[mark=none, my-orange, thick] 
        table[x=frac,y=sparse_err,col sep=comma] {figures/sparsify/ese.csv}; 
        
        \addplot[mark=none, my-purple, thick] 
        table[x=frac,y=sparse_err,col sep=comma] {figures/sparsify/dpp.csv};

        \addlegendentry{\scriptsize{UPR}}
        \addlegendentry{\scriptsize{ESE}}
        \addlegendentry{\scriptsize{DPP}}
    \end{axis}
\end{tikzpicture}

%% file: sections/conclusion.tex
To summarize: we investigated the problem of posterior estimation for dense regression on the exemplary task of depth estimation from light fields.
We therefore contributed the first light field dataset with multimodal depth ground truth.
Additionally, we introduced and compared novel approaches for multimodal light field depth estimation, building on common uncertainty quantification tools.
We observe that methods assuming a single valid depth work well if this assumption holds.
DPP, which predicts arbitrary posterior distributions, works best in general.
Our ESE method does not achieve the same performance, but estimates accurate confidence measures even for wide-baseline light fields.
We hope that our insights lay the foundations for a new line of depth estimation research that overcomes some long-standing limitations of the field.

%% file: sections/acknowledgement.tex
We thank the Center for Information Services and High Performance Computing (ZIH) at TU Dresden for generous allocations of computer time.

%% file: sections/appendix.tex
\vspace{7mm}
\asection{Implementation Details}\label{sec:architectures}
\begin{table}[t]{}
    \begin{center}
        \begin{tabular}{l | r}
            \textbf{Method} &
            \textbf{Parameters} \\
            \hline
            
            EPI-Net &
            $4612166$ \\
            
            UPR &
            $4613300$ \\
            
            ESE &
            $4613300$ \\
            
            DPP &
            $4778872$ \\
            
            \hline
        \end{tabular}
    \end{center}
    \caption{
        Number of \textbf{trainable parameters} for different models
    }
    \label{tab:architecture_n_params}
\end{table}
The architecture of all models in this paper is based on EPI-Net~\cite{shin2018epinet}.
We input four light field view stacks: horizontal, vertical and two diagonals.
Each stack is processed by a separate input stream network.
The horizontal and vertical stacks behave similar when one is rotated by $90^\circ$.
Therefore we effectively share the weights between those two input streams by applying this rotation to the vertical input and revert it before concatenation.
Analogously, we also share weights between the two diagonal input streams.
Subsequently, we concatenate the inferred features, and feed them to an output stream.
All models and streams share the same basic building block which consists of two convolutions with a kernel size of $2 \times 2$.
We use an alternating padding of one and zero and a stride of one to maintain the image dimensions.
In addition, we apply a Rectified Linear Unit (ReLU) non-linearity after the first convolution and a Batch Normalization (BN) as well as a ReLU layer after the second convolution.
\Cref{tab:architecture_n_params} shows the total number of trainable parameters for each model.
A small difference between the four methods is caused by the variable number of output channels.
In the following sections, we describe details, specific to one of the architectures.

\addtolength{\tabcolsep}{\columnSpace}
\begin{table}[t]{}
    \begin{center}
        \begin{tabular}{l | r}
            \textbf{Layer} &
            \textbf{Output Size} \\
        \hline
        LF Stack &
        $B \times 27 \times H \times W$ \\
        \hline
        $2 \times 2$ Conv &
        $B \times 70 \times H \times W$ \\
        ReLU &
        \\
        $2 \times 2$ Conv &
        $B \times 70 \times H \times W$ \\
        BatchNorm &
        \\
        ReLU &
        \\
        \hline
        \multicolumn{2}{c}{Repeat Block $(2 \times)$} \\
        \hline
        \end{tabular}
    \end{center}
    \caption{
        \textbf{Input stream} of EPI-Net, UPR, ESE and DPP\vspace{5mm}
    }
    \label{tab:architecture_epinet_in}
\end{table}
All four methods, share the same backbone network.
The only differences are the variable number of output channels and one additional output ReLU-layer for DPP.
\Cref{tab:architecture_epinet_in} shows the detailed architecture for one input stream.
This subnetwork infers features from one light field stack containing nine images with three color channels, thus a total number of $9 \times 3 = 27$ input channels.
Each input stream consists of three basic blocks.
Because the architecture is based on~\cite{shin2018epinet}, we chose the same number of $70$ output channels.
The features of all input channels are concatenated to a total number of $4 * 70 = 280$ feature channels and fed to the output stream which is illustrated in \cref{tab:architecture_epinet_out}.

\begin{table}[t]{}
    \begin{center}
        \begin{tabular}{l| r}
            \textbf{Layer} &
            O\textbf{utput Size} \\
        \hline
        Concatenate &
        $B \times 280 \times H \times W$ \\
        \hline
        $2 \times 2$ Conv &
        $B \times 280 \times H \times W$ \\
        ReLU &
        \\
        $2 \times 2$ Conv &
        $B \times 280 \times H \times W$ \\
        BatchNorm &
        \\
        ReLU &
        \\
        \hline
        \multicolumn{2}{c}{Repeat Block $(6 \times)$} \\
        \hline
        $2 \times 2$ Conv &
        $B \times C_{\mathrm{out}} \times H \times W$ \\
        ReLU &
        \\
        $2 \times 2$ Conv &
        $B \times C_{\mathrm{out}} \times H \times W$ \\
        (ReLU) &
        \\
        \hline
        \end{tabular}
    \end{center}
    \caption{
        \textbf{Output stream} of EPI-Net, UPR, ESE and DPP
    }
    \label{tab:architecture_epinet_out}
\end{table}
The feed-forward output stream consists of a total number of eight blocks.
Both convolutional layers for each block, except the last, output $280$  channels.
The last block outputs $C_{\mathrm{out}}$ channels, depending on the specific model.
In case of our baseline, $C_{\mathrm{out}} = 1$, because it directly predicts the disparity for each pixel.
For Laplacian distribution prediction, we added a second output channel to also predict $b$, thus $C_{\mathrm{out}} = 2$ for UPR and ESE.
The number of discrete disparity ``classes'', predicted by DPP, can be chosen arbitrarily.
Specifically, we chose $C_{\mathrm{out}} = 108$, thus $108$ ``classes'', motivated by the common BadPix007 metric.

\vspace{5mm}
\subsection{Sub-Pixel EPI-Shift}\label{sec:epi-shift}
Our ESE model utilizes the EPI-Shift transformation, introduced by~\cite{leistner2019learning}.
This shear transformation allows us to apply a disparity offset $\Delta y$ to any light field $x$.
We index the 4D light field in horizontal views $U$, vertical views $V$, image width $W$ and image height $H$ as $x_{uvst}$ ($u = 1 \dots U$, $v = 1 \dots V$, $s = 1 \dots W$, $t = 1 \dots H$).
In contrast to the original method which only applies integer pixel shifts, we also need sub-pixel shifts to ensure the detection of modes that are closer than one pixel.
To achieve this, we apply a linear interpolation.
Thus the original formulation for a horizontal EPI
\begin{equation}
    \shift(x_{uvst}, \Delta y) = x_{uv(s - \Delta y \cdot u)t}
\end{equation}
can be generalized to continuous $\Delta y$ using linear interpolation
\begin{equation}
    \shift(x_{uvst}, \Delta y) = \alpha x_{uv(\lfloor s - \Delta y \cdot u \rfloor)t} + (1 - \alpha) x_{uv(\lceil s - \Delta y \cdot u \rceil)t} 
\end{equation}
with an interpolation factor $\alpha = \mathrm{frac}(\Delta y \cdot u)$.
This can be adapted trivially to vertical EPIs.
For diagonal EPIs, the horizontal and vertical shift is applied successively.

\vspace{5mm}
\asection{Bayesian Interpretation of Opacity}\label{sec:opacity_probability}

From a Bayesian perspective, the probability $p(y_{ij})$ of each possible ground truth disparity value for a pixel
quantifies the ``degree of belief'' in this value.
For a synthetic dataset, in absence of a real ground truth measurement device whose characteristics we can analyze,
any definition for $p(y_{ij})$ is valid as long as it leads to stable training and a model that reproduces
the different modes with their corresponding probabilities faithfully at test time (as we verify in \cref{sec:experiments}).

However, there are still some choices which are more sensible or well founded than others.
In terms of the opacity $\eta_j$, it should be evident to chose
\begin{align}
    \eta_j = 0 &\implies p(y_{ij}) = 0\\
    \eta_j = 1 &\implies p(y_{ij}) = 1,
\end{align}
meaning that if an object is not visible at all in a pixel, its disparity should not be considered,
and vice versa, if an object is the only one visible in a pixel, its disparity should be the only valid answer.
In between these two points, we argue for the simplest choice of $p(y_{ij}) = \eta_j$.
We note that if a setup requires a different definition of $p(y_{ij})$ (e.g. re-weight to increase the dominant mode, up-weight the foreground mode, etc.),
the posterior can easily be re-weighted at test time, without retraining the model.
This is only possible with methods such as ours that produce a full posterior.

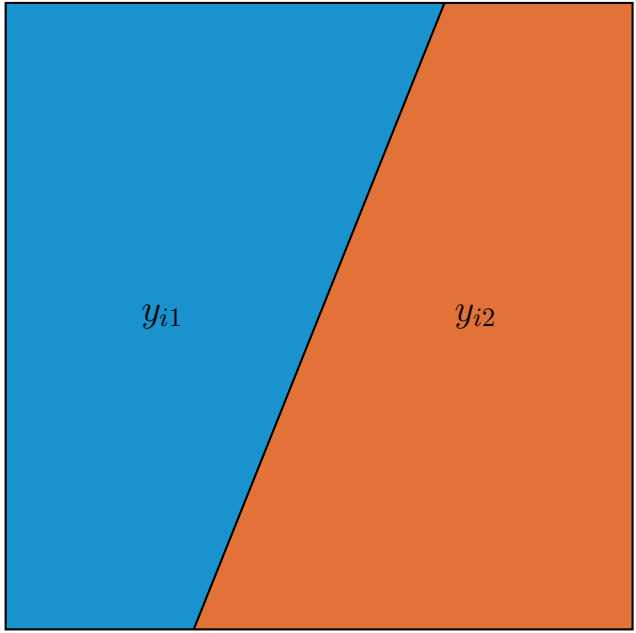
\begin{figure}
    \centering
    \input{figures/square_pixel/square_pixel}
    \caption{View of a single idealized square pixel (synthetic case) containing an edge.
    The opacity values $\eta_{1,2}$ of the rendered pixel correspond to the fraction of the area that the two objects take up within it,
    and therefore to the probability with which the disparity would be measured at a random point in the pixel.
    }
    \label{fig:pixel_opacity_explanation}
\end{figure}

Despite various valid choices of defining $p(y_{ij})$, we do argue that our definition makes practical sense: 
the opacity corresponds to the fraction of the area that an object takes up within in a pixel before integration or rendering.
It is therefore equal to the probability that the depth of that object would be observed when measuring at a random subpixel position.
In other words, if we were to take many physical depth measurements within a pixel,
the relative occurrence of each measured depth value $y_{ij}$ (therefore arguably the probability $p(y_{ij})$), 
would be the same as the opacity $\eta_j$. This is illustrated further in \cref{fig:pixel_opacity_explanation}.
While this applies exactly to our synthetically rendered dataset,
some additional effects such as point spread functions and non-uniform pixel integration functions would apply for real recorded light fields.
These effects might make the derivation more complex, but do not change the general idea.

\asection{Dataset Generation}\label{sec:dataset_generation}
In the following, we describe the generation of our multimodal light field depth dataset:
To maximize occlusions, we generate relatively deep indoor room scenes with a high number of objects.
From a set of $750$ 3D assets, mainly furniture and accessories, we randomly choose $48$ objects per scene and place them in a non-colliding way on the floor.
In addition, random materials with a random opacity are chosen to increase the number of semi-transparent surfaces.
To maximize the diversity, we also randomly choose one of $750$ tileable textures for the walls, ceiling and floor.
We then render the created scene by separating it into $128$ slices of equal depth, as we observed that this leads to different objects falling into different slices almost always.
We then render the color, alpha transparency and depth of each pixel for each slice.
Alpha compositing follows the ``over operator''
\begin{equation}
    C_0 =
    \frac{C_1 \alpha_1 + C_2 \alpha_2 (1 - \alpha_1)}
    {\alpha_0}
\end{equation}
with $C_0$ being the resulting color from color $C_1$ rendered \textit{over} color $C_2$.
The new alpha opacity of color $C_0$ is
\begin{equation}
    \alpha_0 = {\alpha_1 + \alpha_2(1 - \alpha_1)}.
\end{equation}
The contribution $p(y_j) = \eta_j$ of the color $C_j$ at disparity $y_j$ is therefore calculated as
\begin{equation}
    p(y_j) = \eta_j = \alpha_j \left( 1 - \alpha_{j - 1} \left( 1 - \alpha_{j - 2} \left( 1 - \dots \alpha_0  \right)   \right)  \right).
\end{equation}
Lastly, we save all depths for each pixel that are not fully occluded by slices in front.
Note that, apart from the multimodal depth ground truth, our synthetic light fields are similar to real light field recordings.
The multi-layer color information that real light field cameras could not record is not used as an input to our methods.

\asection{Additional Experiments}\label{sec:additional_experiments}
\begin{table*}[h]
    \begin{center}
        \resizebox{\linewidth}{!}{
        \begin{tabular}{l|r r | r r r | r | r}
        \hline
            \textbf{Method} &
            \multicolumn{2}{c|}{\textbf{Unimodal Metrics}} &
            \multicolumn{3}{c|}{\textbf{KL Divergence}} &
            \textbf{AuSE} $\downarrow$&
            \textbf{Time} $\downarrow$\\
            &
            MSE $\downarrow$&
            BadPix $\downarrow$&
            Unimodal $\downarrow$&
            Multimodal $\downarrow$&
            Overall $\downarrow$&
            &  (in sec)    \\
        \hline
            BASE (multi)  &
            \textbf{0.435} & 0.274 & 4.807 & 8.081 & 6.078 & - & \textbf{0.557} \\
            
            UPR (multi) &
            0.480 & 0.285 & 2.028 & 3.551 & 2.448 & \textbf{0.115} & 0.578 \\
            
            ESE (multi) &
            1.204 & 0.245 & 4.330 & 3.769 & 4.226 & 0.182 & 4.502 \\
            
            DPP (multi) &
            0.608 & \textbf{0.239} & \textbf{1.786} & \textbf{3.193} & \textbf{2.136} & 0.288 & 1.068 \\
        \hline
            IBR \cite{sinha2012image} &
            1.436 & 0.365 & 3.835 & 3.436 & 3.843 & 0.617 & 11.263 \\ 
            
            SLFC \cite{johannsen2016sparse}  &
            3.449 & 0.660 & 3.694 & 3.908 & 3.715 & 0.324 & 1054.231 \\
        \hline
        \end{tabular}
        }
    \end{center}
    \vspace{-3mm}
    \caption{
        \textbf{Comparison to IBR~\cite{sinha2012image} and SLFC~\cite{johannsen2016sparse}}, from left to right:
        Mean Squared Error and the common BadPix007 score (percentage of pixels with $|y_i - \hat{y}_i| > 0.07$),
        Kullback-Leibler divergence on unimodal, multimodal and all pixels,
        Area under Sparsification Error (AuSE),
        runtime of one forward pass.
        Our methods were trained using the multimodal loss $\mathcal{L}_x^{\mathrm{MM}}$.
        Lower is better
    }
    \label{tab:results_slfc}
\end{table*}
In this section, we first compare our methods to ``Image-Based Rendering for Scenes with Reflections'' (IBR)~\cite{sinha2012image} and ``What Sparse Light Field Coding Reveals about Scene Structure'' (SLFC)~\cite{johannsen2016sparse}.
Secondly, we present visualizations of exemplary depth posteriors predicted by UPR and DPP.
Thirdly, we evaluate our work on the commonly used HCI 4D Light Field Dataset~\cite{honauer2016dataset} and show additional qualitative results.

\subsection{Comparison to IBR~\cite{sinha2012image} and SLFC~\cite{johannsen2016sparse}}

\begin{figure*}[p]
    \centering
    \begin{subfigure}{0.196\linewidth}
        \includegraphics[width=\textwidth]{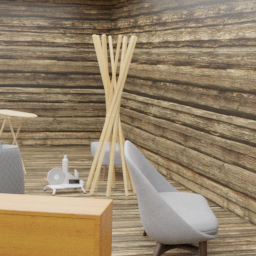}
        \subcaption{Center view}
    \end{subfigure}
    \begin{subfigure}{0.196\linewidth}
        \includegraphics[width=\textwidth]{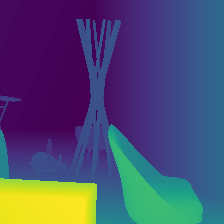}
        \subcaption{Ground Truth}
    \end{subfigure}
    \begin{subfigure}{0.196\linewidth}
        \includegraphics[width=\textwidth]{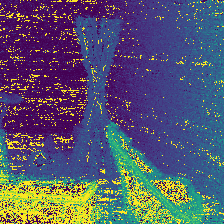}
        \subcaption{SLFC~\cite{johannsen2016sparse}}
    \end{subfigure}
    \begin{subfigure}{0.196\linewidth}
        \includegraphics[width=\textwidth]{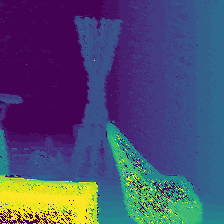}
        \subcaption{IBR~\cite{sinha2012image}}
    \end{subfigure}
    \begin{subfigure}{0.196\linewidth}
        \includegraphics[width=\textwidth]{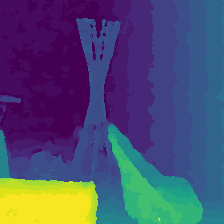}
        \subcaption{DPP}
    \end{subfigure}
    \caption{
        \textbf{Qualitative results of IBR~\cite{sinha2012image} and SLFC~\cite{johannsen2016sparse}, compared to DPP}
        on one of our multimodal validation scenes:
        We chose the disparity which corresponds to the strongest coefficient for each pixel.
        Compared to our deep learning based methods, IBR~\cite{sinha2012image} and SLFC~\cite{johannsen2016sparse} tend to wrong classifications in non-textured areas which causes noise.
        This also has a negative impact on both methods posterior prediction performance.
    \label{fig:results_slfc}
    }
\end{figure*}
We additionally compared our methods to two multimodal depth estimation approaches~\cite{sinha2012image}\cite{johannsen2016sparse}.
These are, to the best of our knowledge, the only previous methods which are able to estimate multiple depth modes.
For ``Image-Based Rendering for Scenes with Reflections''~\cite{sinha2012image} we implemented the normalized cross-correlation framework for our own dataset, as no source code was publicly available.
The method computes the pairwise normalized cross-correlation in a small window $(3\mathrm{px} \times 3\mathrm{px})$ and utilizes it to form a cost volume.
In a second step, up to two disparity planes are extracted from the volume using a modified semi-global matching algorithm.
We interpret the per-pixel normalized cross-correlations as our disparity posterior distributions.
To achieve better results, we first subtract the per-pixel minimum cross-correlation and then normalize the distribution.

We also compared our methods to ``What Sparse Light Field Coding Reveals about Scene Structure'' (SLFC)~\cite{johannsen2016sparse}.
The method uses a dictionary of small EPI-Patches.
Each atom in this dictionary corresponds to a unique disparity.
On small EPI windows around each pixel, the Lasso-Optimizer is used to infer the coefficients for each atom.
A large coefficient for an atom means that the disparity which corresponds to this atom was observed at this pixel.
The vector of coefficients can therefore also be interpreted as a discrete disparity posterior distribution, similarly to DPP.
The authors were able to provide us with only a part of the code which we used to create the dictionaries for our multimodal validation dataset.
We used the Lasso optimizer from the Python framework ``scikit-learn'' and set $\alpha = 0.01$ as recommended by the paper authors.
Finally, we optimized the posterior distribution for each pixel.

We compared both methods to our four deep learning based models.
Please note, that due to the enormous runtime of SLFC (even with our parallel implementation on $128$ CPU cores), we run it on a cropped down $(0.5 \times 0.5)$ version of our validation dataset.
For a fair comparison, we ran all methods trained on the multimodal posterior distribution with loss functions $\mathcal{L}_x^{MM}$ on the same cropped down scenes and chose a the same number of $108$ disparity steps for all methods.

\Cref{tab:results_slfc} shows the results of our comparison.
We notice that IBR and SLFC produce more wrong classifications in non-textured and therefore uncertain areas which leads to more overall noise.
We argue that this is due to the local per-pixel optimization.
In contrast, our neural networks benefit from a larger receptive field and are therefore capable to deliver smooth results, even within relatively large non-textured areas (compare~\cref{fig:results_slfc}).
This effect causes an overall worse performance of IBR and SLFC.
To compute the unimodal metrics, we chose the discrete disparity with the highest posterior probability for each pixel.
Both, the MSE and BadPix score confirm our observations.
Note that IBR and SLFC both perform better than our baseline model in terms of multimodal posterior prediction.
This clearly shows that the methods are indeed able to correctly predict multiple disparity modes.
However, the predicted posterior distributions also suffer from poor performance in uncertain regions.
Additionally, due to each pixel being optimized separately, the runtime of SLFC is several orders of magnitudes higher.
One $256\mathrm{px} \times 256\mathrm{px}$ scene took approximately $18$ minutes to compute in parallel on a dual CPU machine with 128 cores, while DPP runs in approximately one second on a single GPU.

\clearpage
\onecolumn
\subsection{Visualization of Disparity Posterior Distributions}\label{sec:posterior_vis}
\begin{figure*}[h]
    \begin{subfigure}{\textwidth}
        \begin{minipage}[t]{0.24\linewidth}
            \centering
            \vspace{0pt}
            \includegraphics[width=\textwidth]{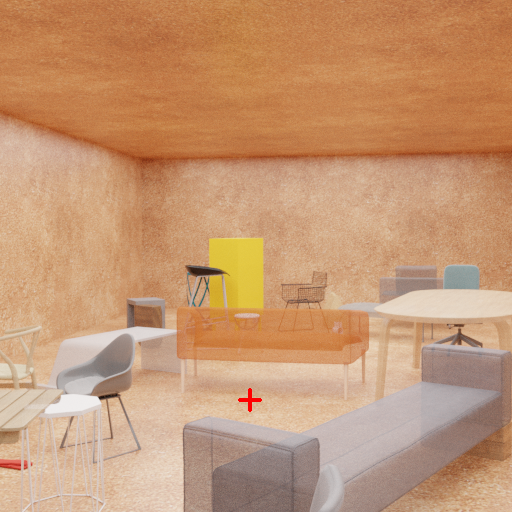}
        \end{minipage}%
        \hfill
        \begin{minipage}[t]{0.74\linewidth}
            \centering
            \vspace{0pt}
            \input{figures/results_posteriors/0013_uni/posteriors}
        \end{minipage}%
        \subcaption{\label{fig:posterior_vis_one}Validation scene $4$: pixel (red cross) contains a single disparity mode\vspace{5mm}}
    \end{subfigure}
    \begin{subfigure}{\textwidth}
        \begin{minipage}[t]{0.24\linewidth}
            \centering
            \vspace{0pt}
            \includegraphics[width=\textwidth]{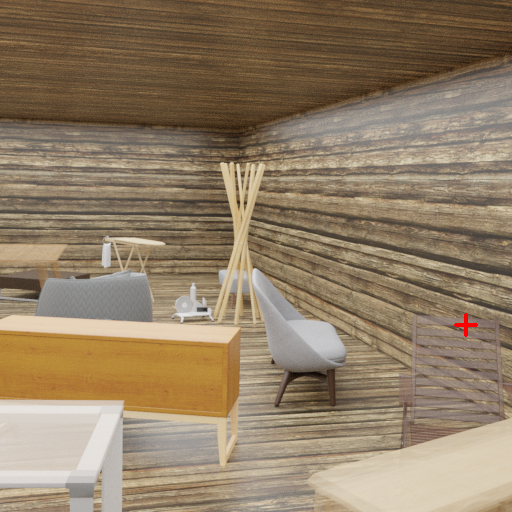}
        \end{minipage}%
        \hfill
        \begin{minipage}[t]{0.74\linewidth}
            \centering
            \vspace{0pt}
            \input{figures/results_posteriors/0007_multi/posteriors}
        \end{minipage}%
        \subcaption{\label{fig:posterior_vis_two}Validation scene $3$: pixel (red cross) contains two disparity modes\vspace{5mm}}
    \end{subfigure}
    \begin{subfigure}{\textwidth}
        \begin{minipage}[t]{0.24\linewidth}
            \centering
            \vspace{0pt}
            \includegraphics[width=\textwidth]{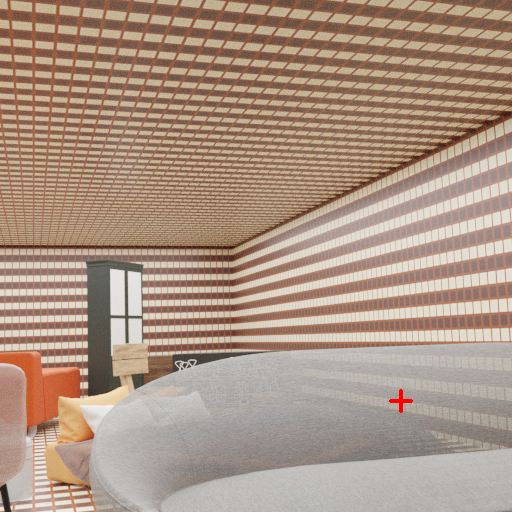}
        \end{minipage}%
        \hfill
        \begin{minipage}[t]{0.74\linewidth}
            \centering
            \vspace{0pt}
            \input{figures/results_posteriors/0123_multi/posteriors}
        \end{minipage}%
        \subcaption{\label{fig:posterior_vis_three}Validation scene $9$: pixel (red cross) contains three disparity modes\vspace{5mm}}
    \end{subfigure}
    \vspace{-3mm}
    \caption{\label{fig:posterior_vis}
    \textbf{Visualization of disparity posterior distributions} for one pixel (red cross) estimated by UPR (orange) and DPP (green) and discrete ground truth posterior (blue).
    Note that DPP is able to estimate up to two modes reliably, while UPR only picks up a single mode.
    }
\end{figure*}

To give some examples of predicted posterior distributions, we visualized estimations of UPR and DPP on our multimodal validation dataset.
Similar to our evaluations, we discretized the ground truth disparity posterior using the same number of bins.
We chose certain pixels from three validation scenes that contain one, two and three disparity modes respectively.
Note that DPP manages to detect both modes in~\cref{fig:posterior_vis_two}, but outputs a high uncertainty due to the similar colors of the foreground and background object.
In~\cref{fig:posterior_vis_three}, two of the tree modes collapsed into one.
UPR always picked up one present mode with a high uncertainty.
Please note that our dataset randomly adds transparency to object materials.
This causes some objects that would be opaque in real life to become transparent.
\vspace{2pt}

\clearpage

\subsection{Evaluation on HCI 4D Light Field Dataset~\cite{honauer2016dataset}}
We also evaluated our methods on the commonly used HCI 4D Light Field Dataset~\cite{honauer2016dataset}.
Like previous methods~\cite{shin2018epinet}, we used the 16 ``additional'' scenes as our training dataset and the four ``training'' scenes for validation.
As this dataset only contains a single ground truth depth, we used the unimodal loss functions $\mathcal{L}_x$.
All other training parameters remain the same as mentioned in~\cref{sec:experiments}.
Note, that we only trained on the HCI dataset for this particular experiment.
The methods used in all other experiments were trained solely on our novel multimodal dataset.

\begin{figure*}[h]
    \centering
    \begin{subfigure}{0.495\textwidth}
        \resizebox{\linewidth}{!}{\input{figures/sparsify_hci/cnn.tex}}
        \subcaption{Sparsification curve}
        \label{fig:sparsify_curve}
    \end{subfigure}
    \begin{subfigure}{0.495\textwidth}
        \resizebox{\linewidth}{!}{\input{figures/sparsify_hci/error.tex}}
        \subcaption{Sparsification Error for all methods\label{fig:sparsify_error}}
    \end{subfigure}
    \caption{
        \textbf{Unimodal uncertainty quantification on HCI 4D Light Field Dataset:} Sparsification results of analyzed methods with respect to the disparity BadPix007.
    }
    \label{fig:sparsify_hci}
\end{figure*}
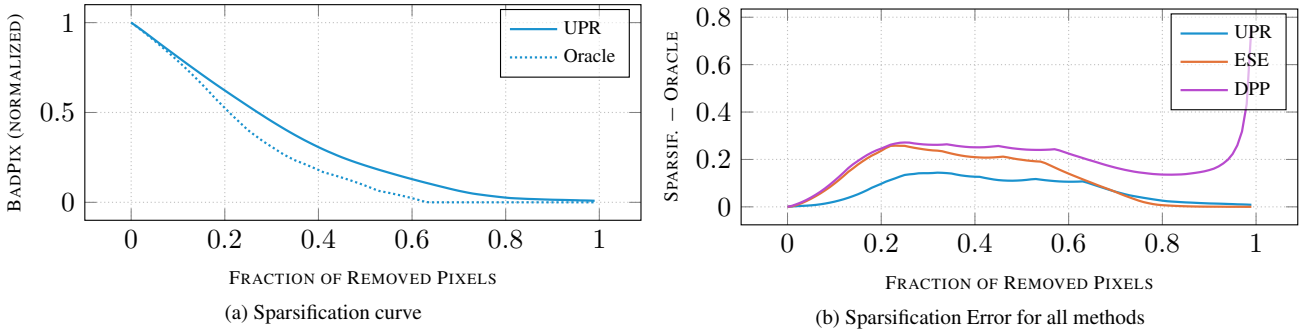

\begin{table}[h]
    \begin{center}
        \begin{tabular}{l|r r | r | r}
        \hline
            \textbf{Method} &
            \multicolumn{2}{c|}{\textbf{Unimodal Metrics}} &
            \textbf{AuSE} $\downarrow$&
            \textbf{Time} $\downarrow$\\
            &
            MSE $\downarrow$&
            BadPix $\downarrow$&
            &  (in sec)    \\
        \hline
        \hline
            BASE &
            \textbf{0.011} & 0.065 & - & \textbf{0.480} \\
            
            UPR &
            0.012 & 0.056 & \textbf{0.060} & 0.481 \\
            
            ESE &
            0.163 & 0.088 & 0.091 & 14.863 \\
            
            DPP &
            0.018 & \textbf{0.044} & 0.110 & 0.783 \\
            
        \hline
        \end{tabular}
    \end{center}
    \vspace{-3mm}
    \caption{
        \textbf{Evaluation on HCI dataset~\cite{honauer2016dataset}}, from left to right:
        Mean Squared Error and the common BadPix007 score (percentage of pixels with $|y_i - \hat{y}_i| > 0.07$),
        Area under Sparsification Error (AuSE),
        runtime of one forward pass.
        Lower is better
    }
    \label{tab:results_hci}
\end{table}

\begin{figure*}[p]
    \centering
    \begin{subfigure}{0.4449\linewidth}
        \includegraphics[width=\textwidth]{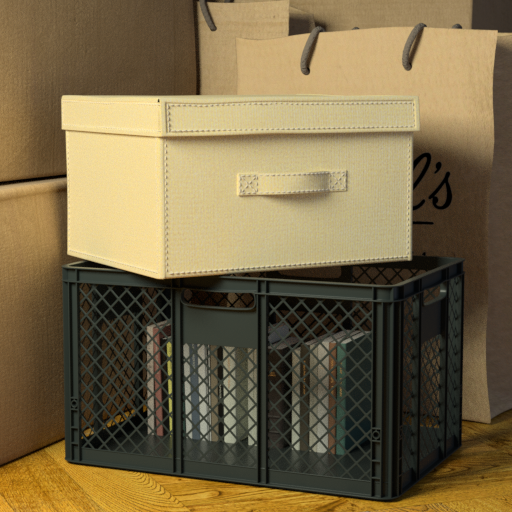}
        \subcaption{Light field \vspace{3mm}}
    \end{subfigure}
    \begin{subfigure}{0.4449\linewidth}
        \includegraphics[width=\textwidth]{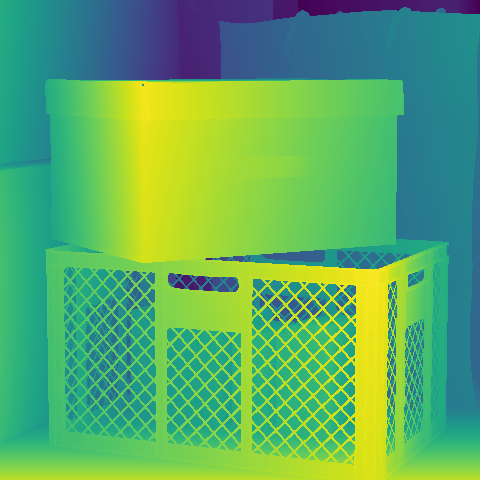}
        \subcaption{Dataset ground truth \vspace{3mm}}
    \end{subfigure}
    \begin{subfigure}{0.22\linewidth}
        \includegraphics[width=\textwidth]{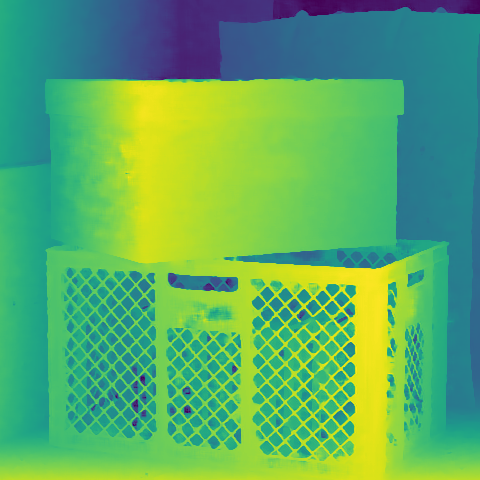}
        \includegraphics[width=\textwidth]{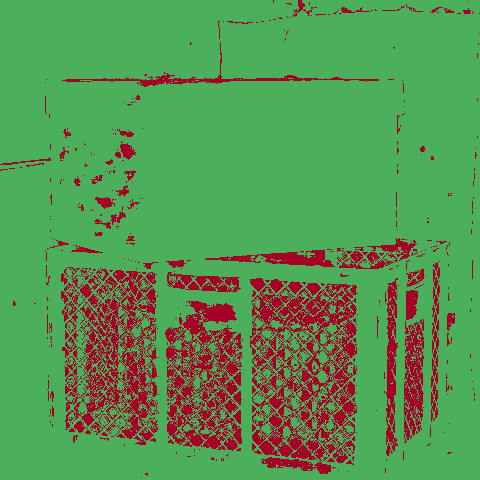}
        \includegraphics[width=\textwidth]{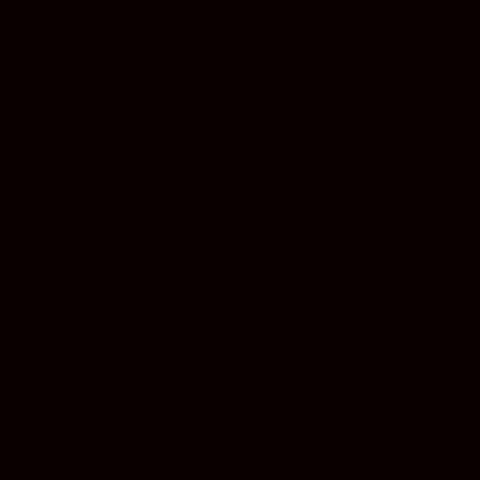}
        \subcaption{BASE}
    \end{subfigure}
    \begin{subfigure}{0.22\linewidth}
        \includegraphics[width=\textwidth]{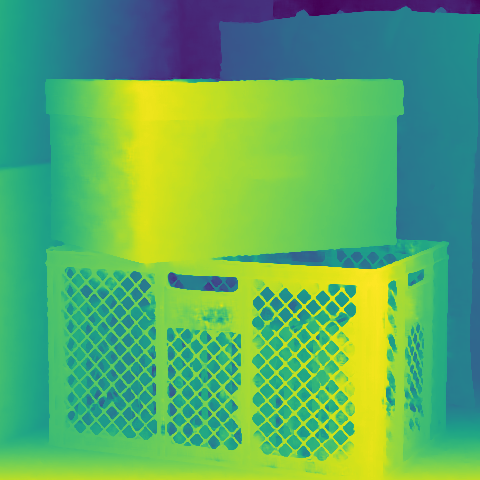}
        \includegraphics[width=\textwidth]{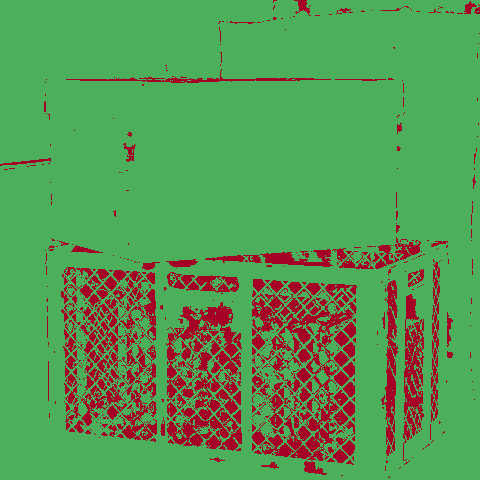}
        \includegraphics[width=\textwidth]{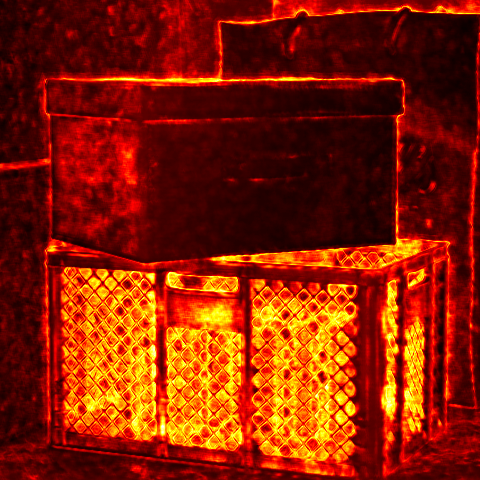}
        \subcaption{UPR}
    \end{subfigure}
    \begin{subfigure}{0.22\linewidth}
        \includegraphics[width=\textwidth]{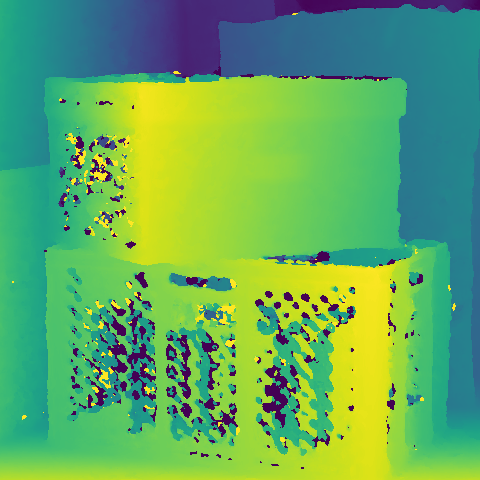}
        \includegraphics[width=\textwidth]{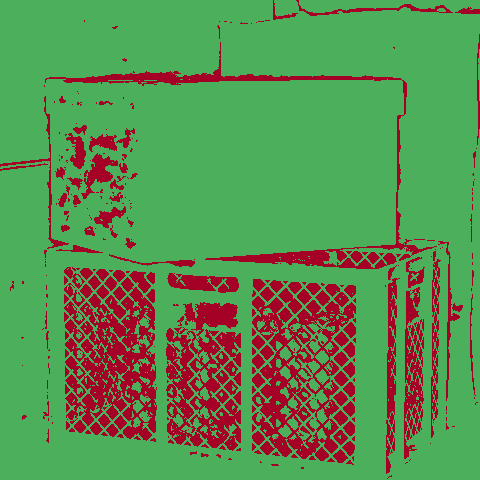}
        \includegraphics[width=\textwidth]{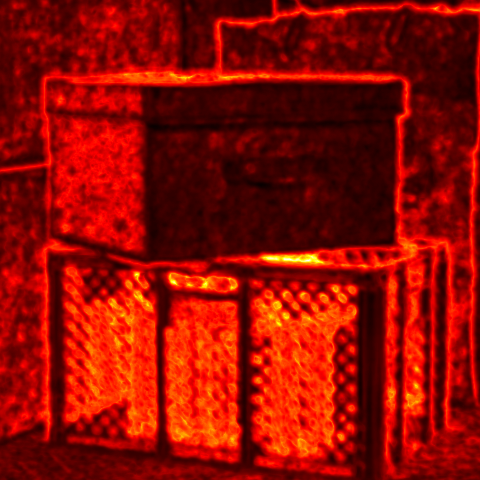}
        \subcaption{ESE}
    \end{subfigure}
    \begin{subfigure}{0.22\linewidth}
        \includegraphics[width=\textwidth]{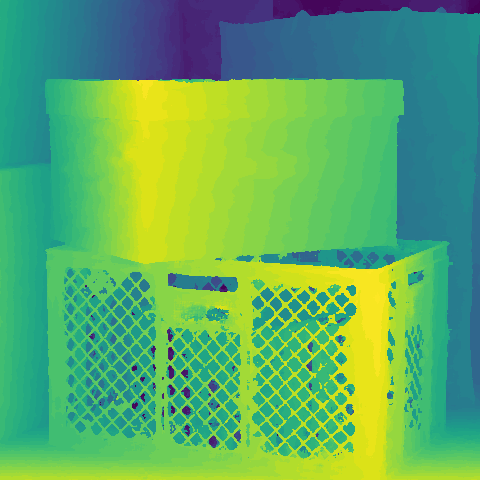}
        \includegraphics[width=\textwidth]{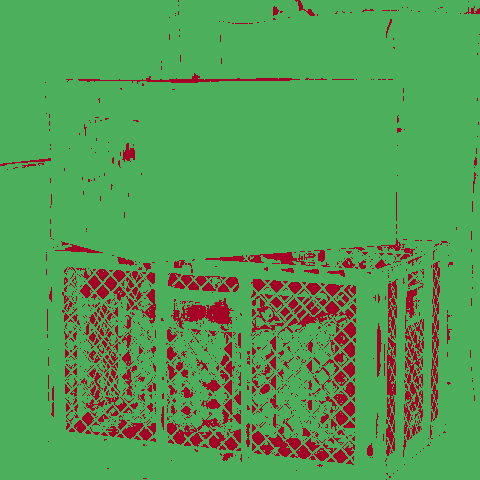}
        \includegraphics[width=\textwidth]{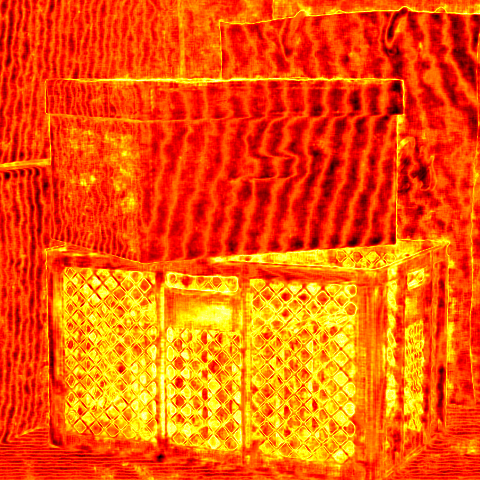}
        \subcaption{DPP}
    \end{subfigure}
    \caption{
        Results of the four posterior prediction methods ((c) - (f)) for \textbf{`boxes'} scene.
        Top: output disparity (most likely mode).
        Center: per-pixel BadPix metric (a pixel $i$ is red if $|y_i - \hat{y}_i| > 0.07$).
        Bottom: per-pixel uncertainty $\sigma^2$ (non-existent for baseline method)
    }
    \label{fig:results_add_boxes}
\end{figure*}

\begin{figure*}[p]
    \centering
    \begin{subfigure}{0.4449\linewidth}
        \includegraphics[width=\textwidth]{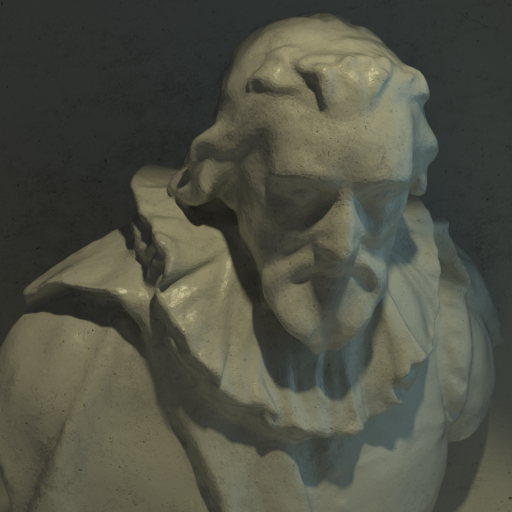}
        \subcaption{Light field \vspace{3mm}}
    \end{subfigure}
    \begin{subfigure}{0.4449\linewidth}
        \includegraphics[width=\textwidth]{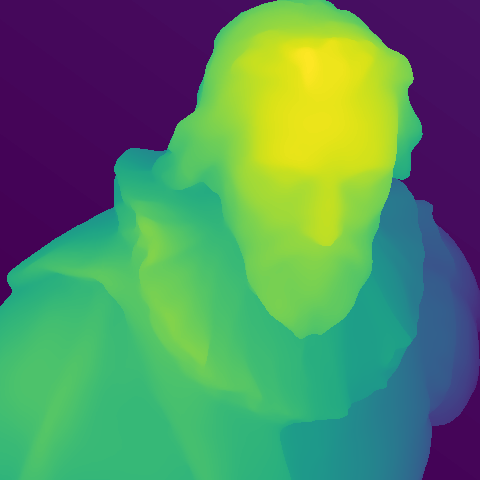}
        \subcaption{Dataset ground truth \vspace{3mm}}
    \end{subfigure}
    \begin{subfigure}{0.22\linewidth}
        \includegraphics[width=\textwidth]{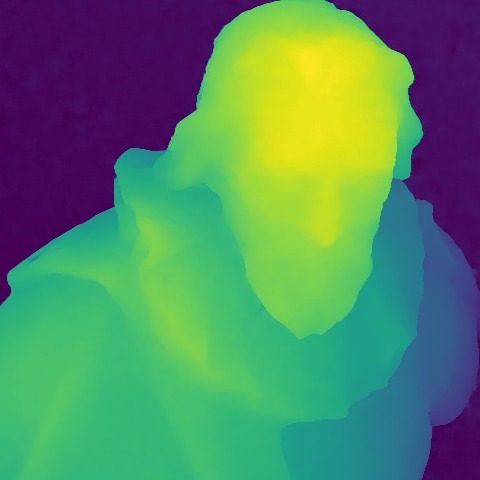}
        \includegraphics[width=\textwidth]{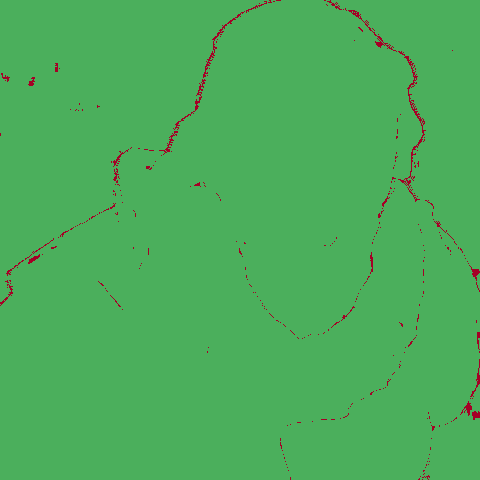}
        \includegraphics[width=\textwidth]{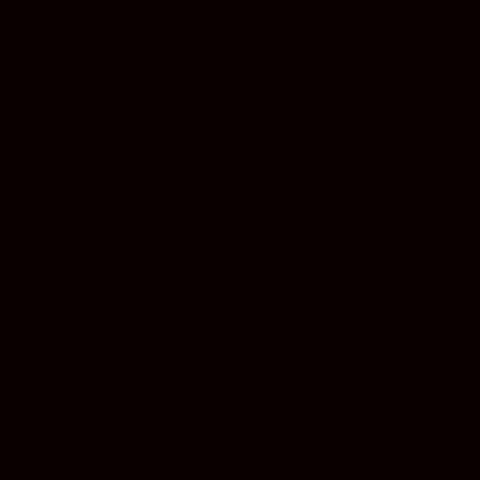}
        \subcaption{BASE}
    \end{subfigure}
    \begin{subfigure}{0.22\linewidth}
        \includegraphics[width=\textwidth]{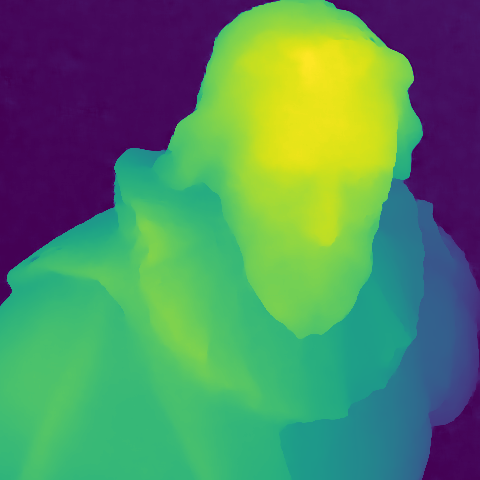}
        \includegraphics[width=\textwidth]{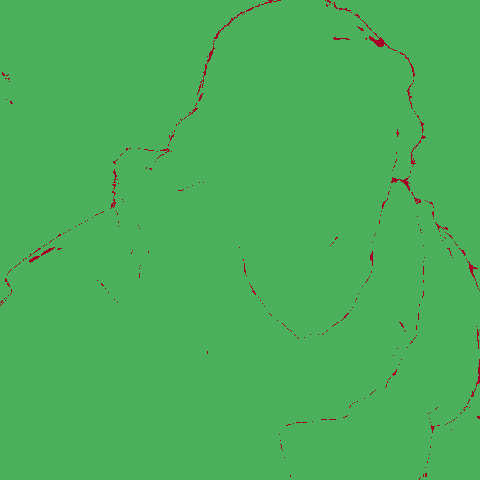}
        \includegraphics[width=\textwidth]{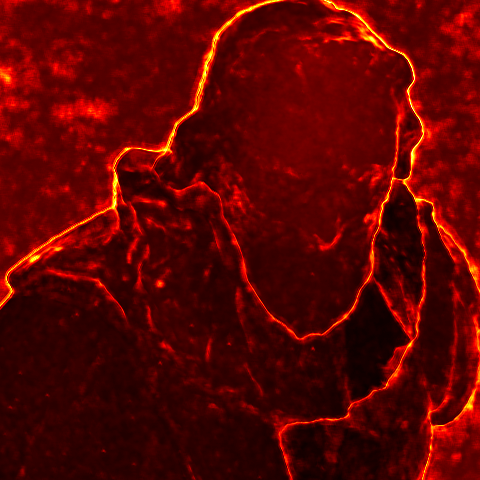}
        \subcaption{UPR}
    \end{subfigure}
    \begin{subfigure}{0.22\linewidth}
        \includegraphics[width=\textwidth]{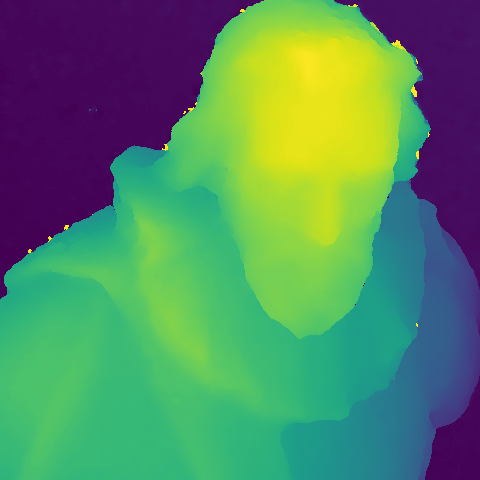}
        \includegraphics[width=\textwidth]{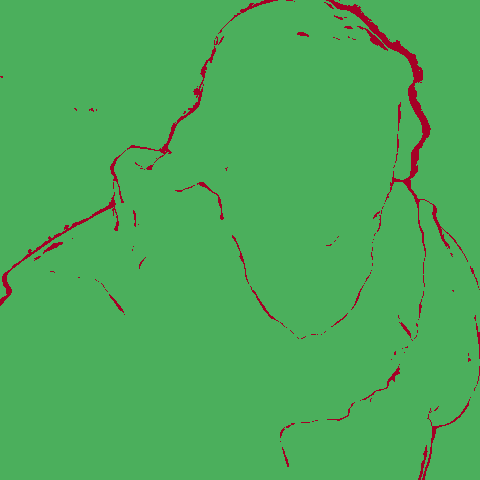}
        \includegraphics[width=\textwidth]{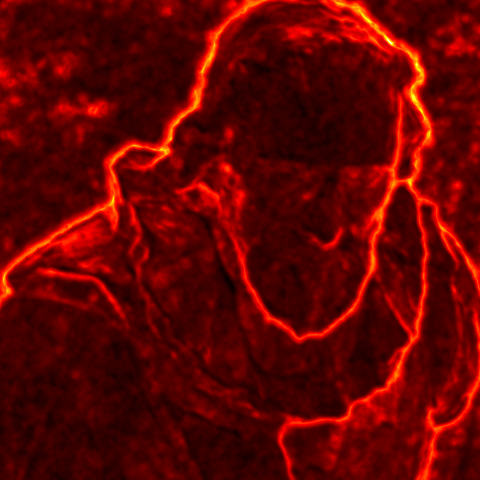}
        \subcaption{ESE}
    \end{subfigure}
    \begin{subfigure}{0.22\linewidth}
        \includegraphics[width=\textwidth]{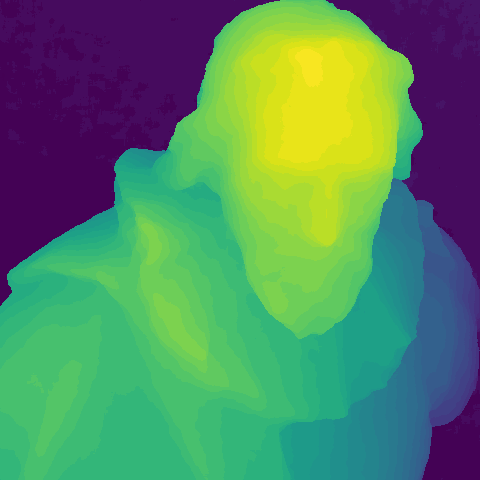}
        \includegraphics[width=\textwidth]{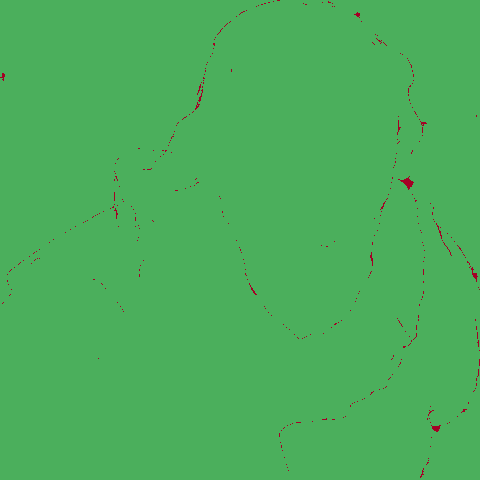}
        \includegraphics[width=\textwidth]{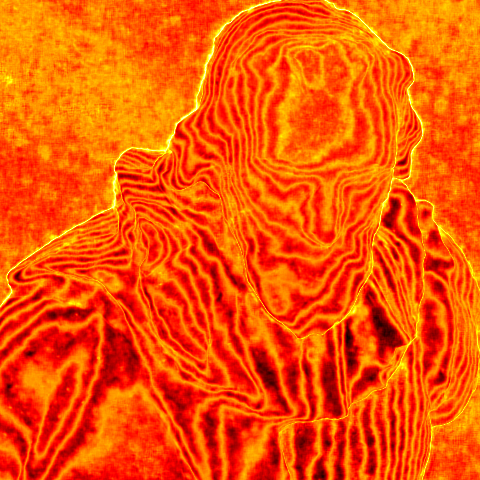}
        \subcaption{DPP}
    \end{subfigure}
    \caption{
        Results of the four posterior prediction methods ((c) - (f)) for \textbf{`cotton'} scene.
        Top: output disparity (most likely mode).
        Center: per-pixel BadPix metric (a pixel $i$ is red if $|y_i - \hat{y}_i| > 0.07$).
        Bottom: per-pixel uncertainty $\sigma^2$ (non-existent for baseline method)
    }
    \label{fig:results_add_cotton}
\end{figure*}

\begin{figure*}[p]
    \centering
    \begin{subfigure}{0.4449\linewidth}
        \includegraphics[width=\textwidth]{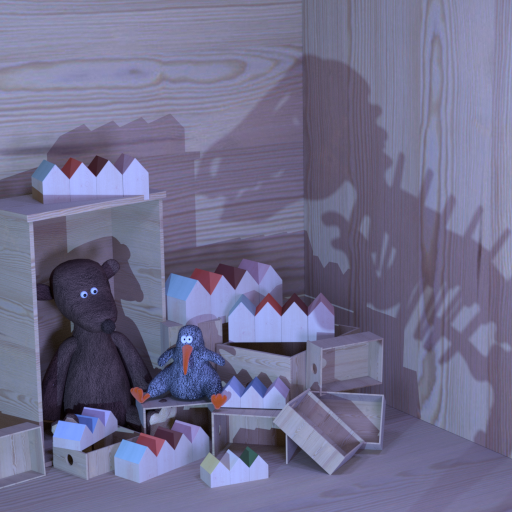}
        \subcaption{Light field \vspace{3mm}}
    \end{subfigure}
    \begin{subfigure}{0.4449\linewidth}
        \includegraphics[width=\textwidth]{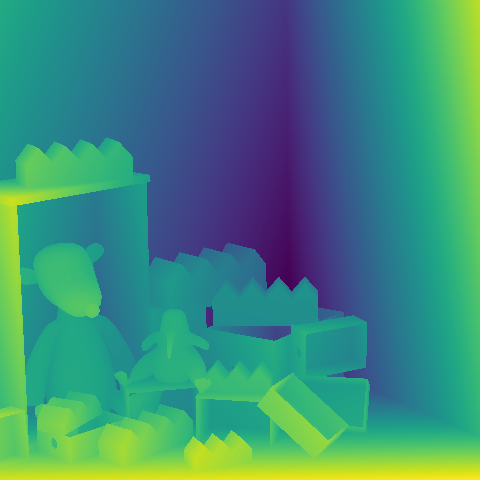}
        \subcaption{Dataset ground truth \vspace{3mm}}
    \end{subfigure}
    \begin{subfigure}{0.22\linewidth}
        \includegraphics[width=\textwidth]{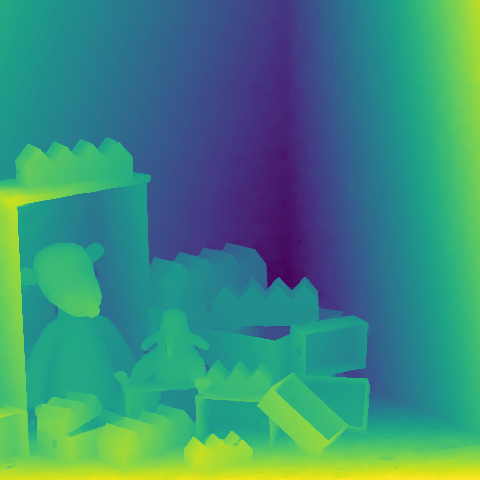}
        \includegraphics[width=\textwidth]{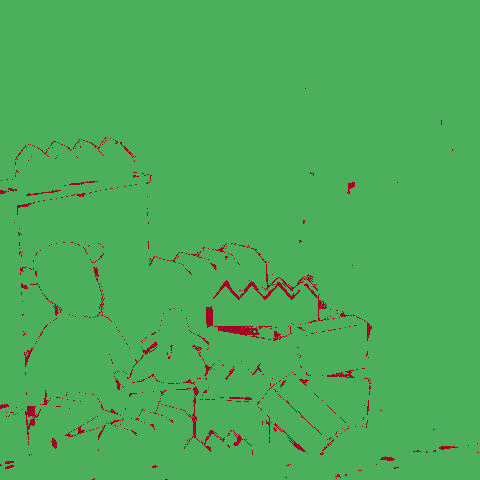}
        \includegraphics[width=\textwidth]{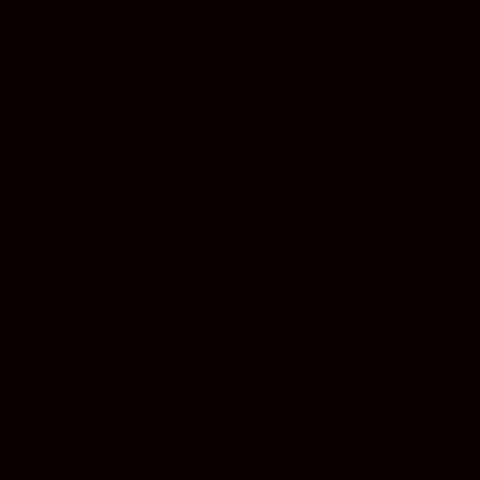}
        \subcaption{BASE}
    \end{subfigure}
    \begin{subfigure}{0.22\linewidth}
        \includegraphics[width=\textwidth]{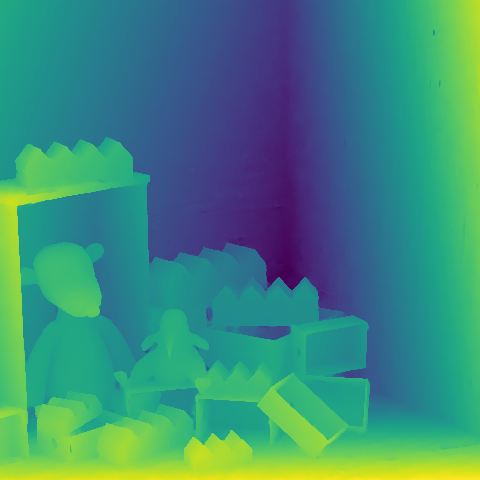}
        \includegraphics[width=\textwidth]{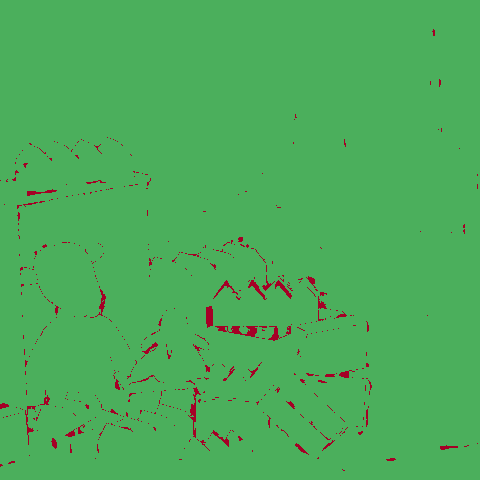}
        \includegraphics[width=\textwidth]{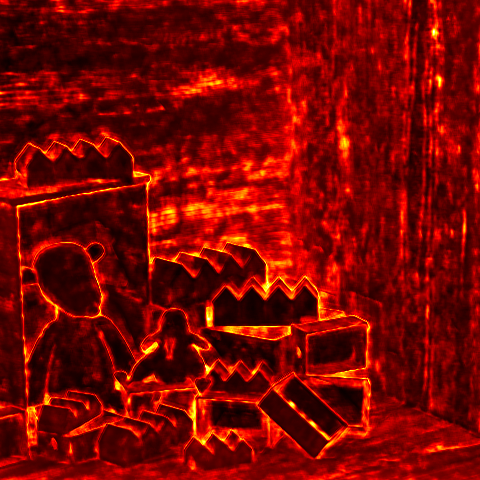}
        \subcaption{UPR}
    \end{subfigure}
    \begin{subfigure}{0.22\linewidth}
        \includegraphics[width=\textwidth]{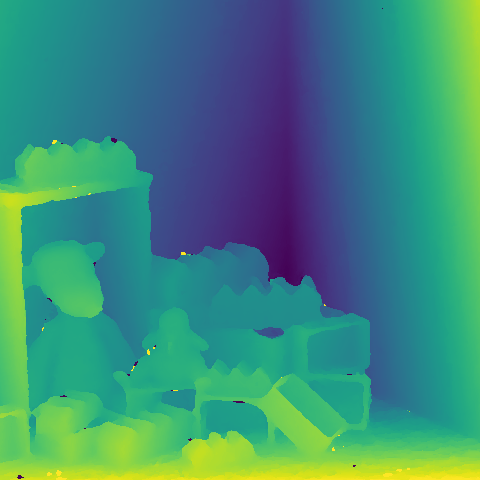}
        \includegraphics[width=\textwidth]{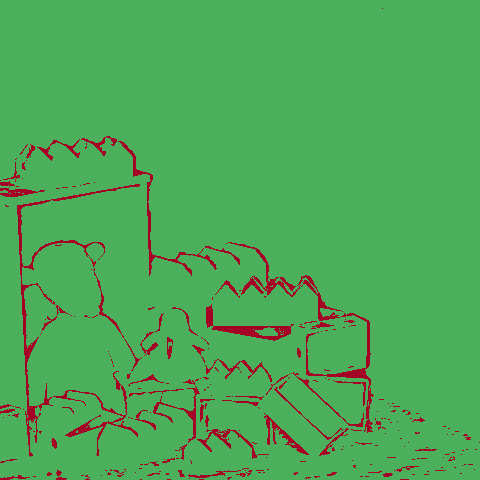}
        \includegraphics[width=\textwidth]{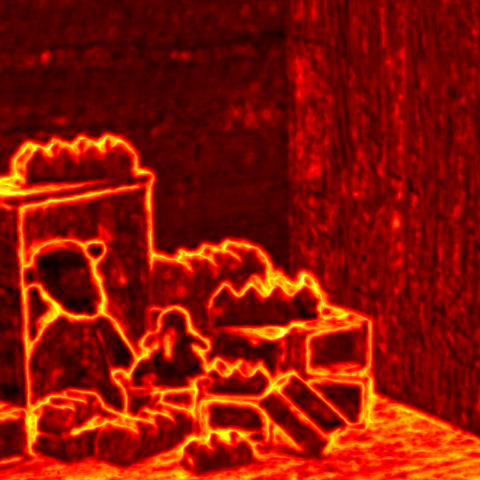}
        \subcaption{ESE}
    \end{subfigure}
    \begin{subfigure}{0.22\linewidth}
        \includegraphics[width=\textwidth]{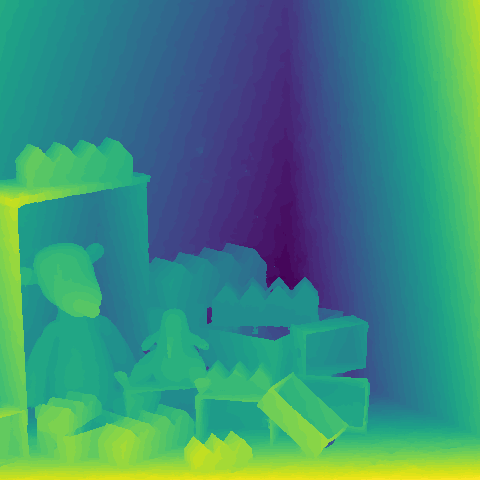}
        \includegraphics[width=\textwidth]{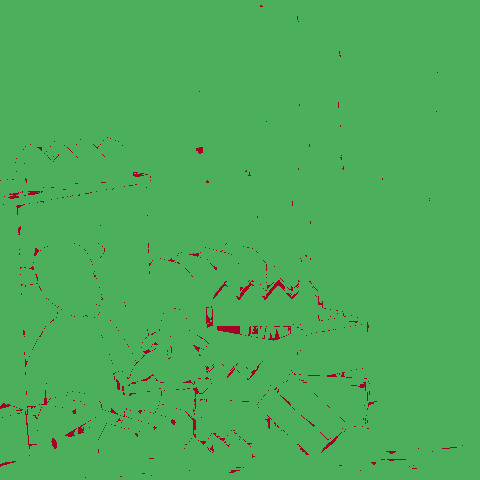}
        \includegraphics[width=\textwidth]{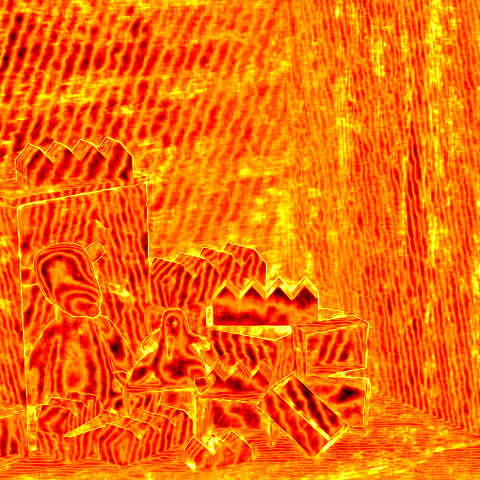}
        \subcaption{DPP}
    \end{subfigure}
    \caption{
        Results of the four posterior prediction methods ((c) - (f)) for \textbf{`dino'} scene.
        Top: output disparity (most likely mode).
        Center: per-pixel BadPix metric (a pixel $i$ is red if $|y_i - \hat{y}_i| > 0.07$).
        Bottom: per-pixel uncertainty $\sigma^2$ (non-existent for baseline method)
    }
    \label{fig:results_add_dino}
\end{figure*}

\begin{figure*}[p]
    \centering
    \begin{subfigure}{0.4449\linewidth}
        \includegraphics[width=\textwidth]{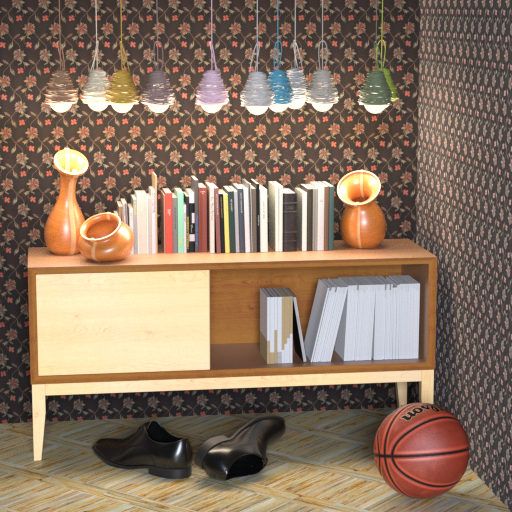}
        \subcaption{Light field \vspace{3mm}}
    \end{subfigure}
    \begin{subfigure}{0.4449\linewidth}
        \includegraphics[width=\textwidth]{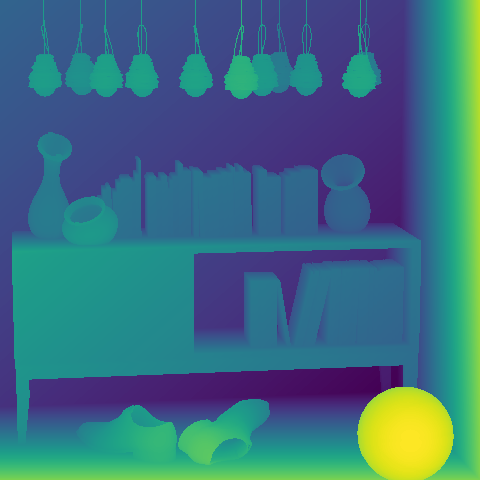}
        \subcaption{Dataset ground truth \vspace{3mm}}
    \end{subfigure}
    \begin{subfigure}{0.22\linewidth}
        \includegraphics[width=\textwidth]{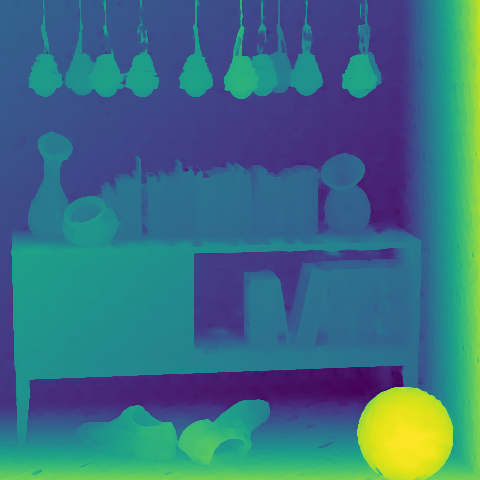}
        \includegraphics[width=\textwidth]{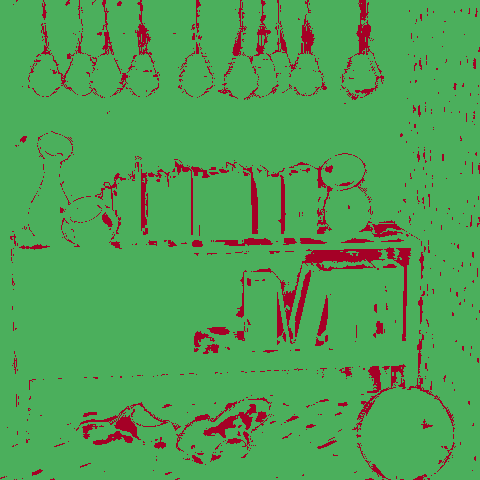}
        \includegraphics[width=\textwidth]{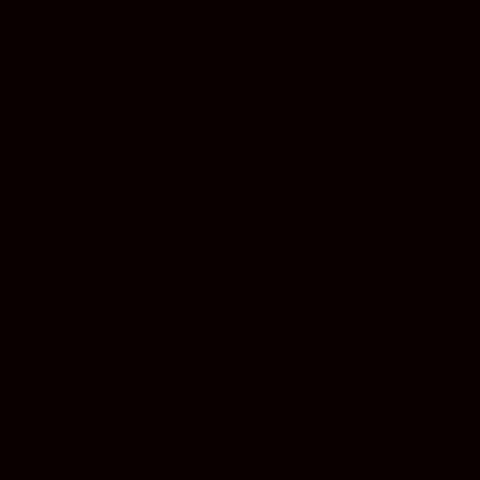}
        \subcaption{BASE}
    \end{subfigure}
    \begin{subfigure}{0.22\linewidth}
        \includegraphics[width=\textwidth]{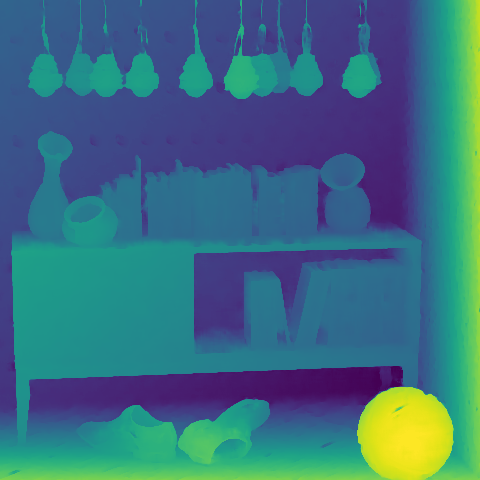}
        \includegraphics[width=\textwidth]{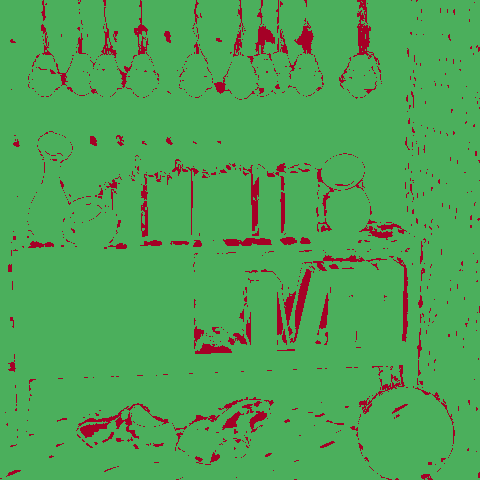}
        \includegraphics[width=\textwidth]{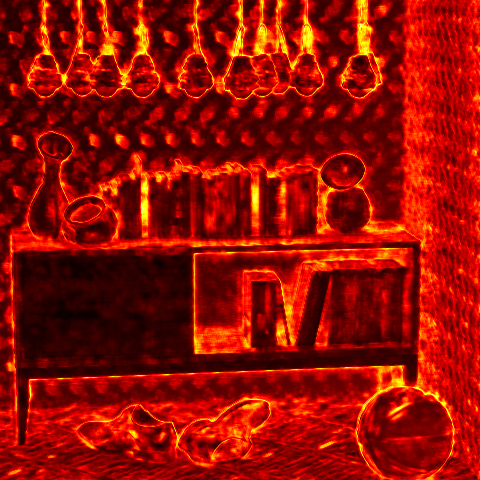}
        \subcaption{UPR}
    \end{subfigure}
    \begin{subfigure}{0.22\linewidth}
        \includegraphics[width=\textwidth]{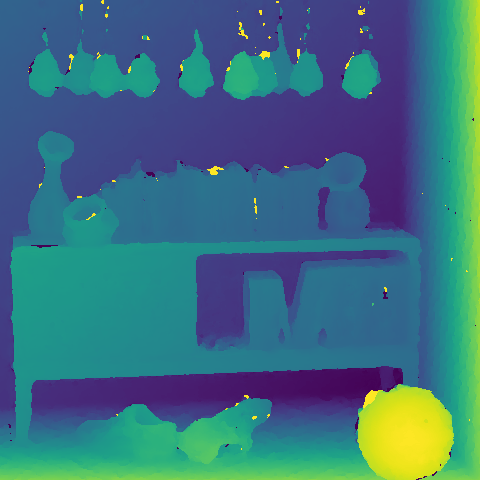}
        \includegraphics[width=\textwidth]{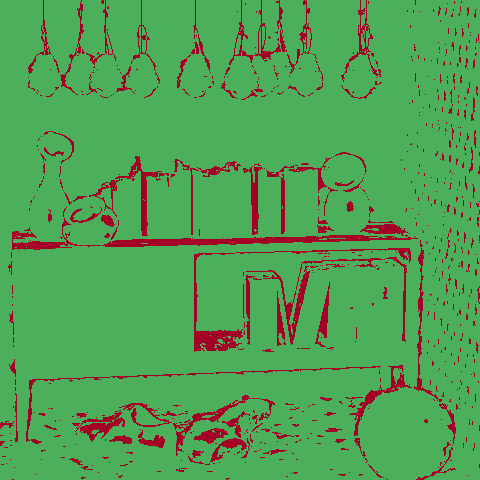}
        \includegraphics[width=\textwidth]{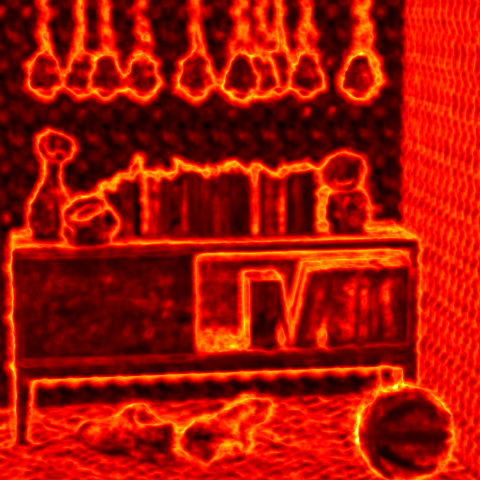}
        \subcaption{ESE}
    \end{subfigure}
    \begin{subfigure}{0.22\linewidth}
        \includegraphics[width=\textwidth]{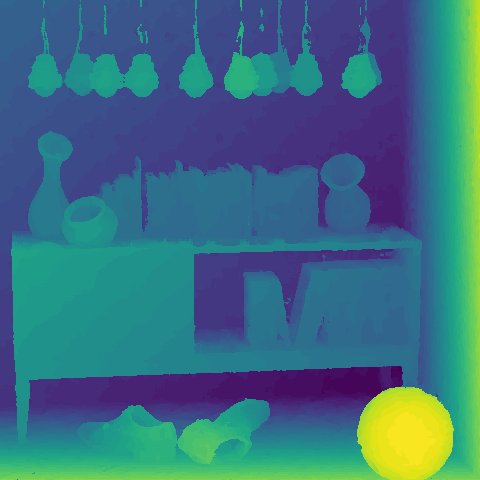}
        \includegraphics[width=\textwidth]{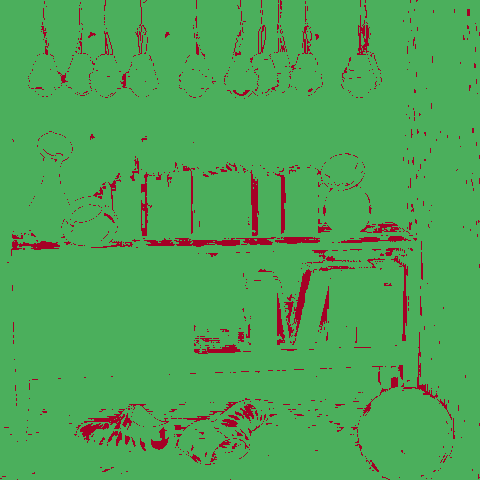}
        \includegraphics[width=\textwidth]{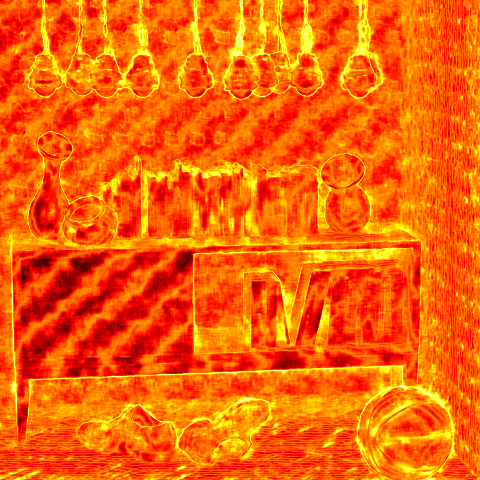}
        \subcaption{DPP}
    \end{subfigure}
    \caption{
        Results of the four posterior prediction methods ((c) - (f)) for \textbf{`sideboard'} scene.
        Top: output disparity (most likely mode).
        Center: per-pixel BadPix metric (a pixel $i$ is red if $|y_i - \hat{y}_i| > 0.07$).
        Bottom: per-pixel uncertainty $\sigma^2$ (non-existent for baseline method)
    }
    \label{fig:results_add_sideboard}
\end{figure*}

\Cref{fig:sparsify_hci} and~\Cref{tab:results_hci} show our experimental results, which are overall very consistent with the experiments on our randomly generated multimodal dataset.
DPP performs best with respect to the amount of accurately predicted pixels (BadPix) but is overconfident which is clearly visible in the sparsification error.
In contrast, UPR and ESE deliver a better sparsification performance.
Qualitative results are shown in~\cref{fig:results_add_boxes} to \cref{fig:results_add_sideboard}.
\vspace{3mm}
\begin{figure*}[h]
    \centering
    \begin{subfigure}{\linewidth}
        \includegraphics[width=\textwidth]{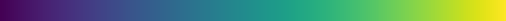}
        $y_{\min}$
        \hfill
        $y_{\max}$
        \vspace{-3mm}
        \subcaption{Disparity \vspace{6mm}}
    \end{subfigure}
    \begin{subfigure}{\linewidth}
        \includegraphics[width=\textwidth]{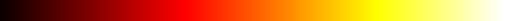}
        $0$
        \hfill
        $\sigma^2_{\max}$
        \vspace{-3mm}
        \subcaption{Uncertainty \vspace{6mm}}
    \end{subfigure}
    \vspace{-2mm}
    \caption{
        \textbf{Color maps} used for results.
        Disparity and uncertainty maps are normalized to enhance visibility
    }
    \label{fig:color_maps}
\end{figure*}

%% file: figures/square_pixel/square_pixel.tex
\begin{tikzpicture}
    \fill[my-orange] (0, 0) rectangle (\linewidth, \linewidth);
    \fill[my-blue] (0, 0) -- (0.3\linewidth, 0) -- (0.7\linewidth, \linewidth) -- (0, \linewidth);
    \draw[thick] (0.3\linewidth, 0) -- (0.7\linewidth, \linewidth);
    \draw[thick] (0, 0) rectangle (\linewidth, \linewidth);
    
    \node at (0.25\linewidth, 0.5\linewidth) {\Large $y_{i1}$};
    \node at (0.75\linewidth, 0.5\linewidth) {\Large $y_{i2}$};
\end{tikzpicture}

%% file: figures/results_posteriors/0013_uni/posteriors.tex
\begin{tikzpicture}
    \begin{axis}[
        width=\linewidth,
        height=0.422\linewidth,
        legend cell align={left},
        grid=major,
        grid style={densely dotted, gray!50},
        legend style={fill=white, fill opacity=1, draw opacity=1,text opacity=1},
        x label style={at={(axis description cs:0.5,0.07)}},
        y label style={at={(axis description cs:0.07,0.5)}},
        ybar interval,
        ylabel=$p(y | x)$,
        xlabel=discretized disparity $y$,
        xmin=1.2,
        xmax=2.2,
        xticklabels={,,},
        yticklabels={,,},
        ]
        
        \addplot[mark=none, my-orange, fill] 
        table[x=y,y=p,col sep=comma] {figures/results_posteriors/0013_uni/upr.csv};
        
        \addplot[mark=none, my-green, fill] 
        table[x=y,y=p,col sep=comma] {figures/results_posteriors/0013_uni/dpp.csv}; 
        
        \addplot[mark=none, my-blue, fill] 
        table[x=y,y=gt,col sep=comma] {figures/results_posteriors/0013_uni/upr.csv}; 
        
        \addlegendentry{\scriptsize{UPR}}
        \addlegendentry{\scriptsize{DPP}}
        \addlegendentry{\scriptsize{GT}}
    \end{axis}
\end{tikzpicture}

%% file: figures/results_posteriors/0007_multi/posteriors.tex
\begin{tikzpicture}
    \begin{axis}[
        width=\linewidth,
        height=0.422\linewidth,
        legend cell align={left},
        grid=major,
        grid style={densely dotted, gray!50},
        legend style={fill=white, fill opacity=1, draw opacity=1,text opacity=1},
        x label style={at={(axis description cs:0.5,0.07)}},
        y label style={at={(axis description cs:0.07,0.5)}},
        ybar interval,
        ylabel=$p(y | x)$,
        xlabel=discretized disparity $y$,
        xmin=1.5,
        xmax=2.5,
        xticklabels={,,},
        yticklabels={,,},
        ]
        
        \addplot[mark=none, my-orange, fill] 
        table[x=y,y=p,col sep=comma] {figures/results_posteriors/0007_multi/upr.csv};
        
        \addplot[mark=none, my-green, fill] 
        table[x=y,y=p,col sep=comma] {figures/results_posteriors/0007_multi/dpp.csv}; 
        
        \addplot[mark=none, my-blue, fill] 
        table[x=y,y=gt,col sep=comma] {figures/results_posteriors/0007_multi/upr.csv}; 
        
        \addlegendentry{\scriptsize{UPR}}
        \addlegendentry{\scriptsize{DPP}}
        \addlegendentry{\scriptsize{GT}}
    \end{axis}
\end{tikzpicture}

%% file: figures/results_posteriors/0123_multi/posteriors.tex
\begin{tikzpicture}
    \begin{axis}[
        width=\linewidth,
        height=0.422\linewidth,
        legend cell align={left},
        grid=major,
        grid style={densely dotted, gray!50},
        legend style={fill=white, fill opacity=1, draw opacity=1,text opacity=1},
        x label style={at={(axis description cs:0.5,0.07)}},
        y label style={at={(axis description cs:0.07,0.5)}},
        ybar interval,
        ylabel=$p(y | x)$,
        xlabel=discretized disparity $y$,
        xmin=1,
        xmax=3.75,
        xticklabels={,,},
        yticklabels={,,},
        ]
        
        \addplot[mark=none, my-orange, fill] 
        table[x=y,y=p,col sep=comma] {figures/results_posteriors/0123_multi/upr.csv};
        
        \addplot[mark=none, my-green, fill] 
        table[x=y,y=p,col sep=comma] {figures/results_posteriors/0123_multi/dpp.csv}; 
        
        \addplot[mark=none, my-blue, fill] 
        table[x=y,y=gt,col sep=comma] {figures/results_posteriors/0123_multi/upr.csv}; 
        
        \addlegendentry{\scriptsize{UPR}}
        \addlegendentry{\scriptsize{DPP}}
        \addlegendentry{\scriptsize{GT}}
    \end{axis}
\end{tikzpicture}

%% file: figures/sparsify_hci/cnn.tex
\begin{tikzpicture}
    \begin{axis}[
        width=\linewidth,
        height=0.5\linewidth,
        legend cell align={left},
        legend style={fill=white, fill opacity=0.6, draw opacity=1,text opacity=1},
        grid=major,
        grid style={densely dotted, gray!50},
        x label style={at={(axis description cs:0.5,0.0)}},
        y label style={at={(axis description cs:0.05,0.5)}},
        ylabel=\scriptsize{\textsc{BadPix (normalized)}},
        xlabel=\scriptsize{\textsc{Fraction of Removed Pixels}},
        ]

        \addplot[mark=none, my-blue, thick]
        table[x=frac,y=uncert,col sep=comma] {figures/sparsify_hci/upr.csv}; 
        
        \addplot[mark=none, my-blue, thick, densely dotted]
        table[x=frac,y=oracle,col sep=comma] {figures/sparsify_hci/upr.csv}; 
        
        \addlegendentry{\scriptsize{UPR}}
        \addlegendentry{\scriptsize{Oracle}}
    \end{axis}
\end{tikzpicture}

%% file: figures/sparsify_hci/error.tex
\begin{tikzpicture}
    \begin{axis}[
        width=\linewidth,
        height=0.5\linewidth,
        legend cell align={left},
        legend style={fill=white, fill opacity=0.6, draw opacity=1,text opacity=1},
        grid=major,
        grid style={densely dotted, gray!50},
        x label style={at={(axis description cs:0.5,0.0)}},
        y label style={at={(axis description cs:0.05,0.5)}},
        ylabel=\scriptsize{\textsc{Sparsif. $-$ Oracle}},
        xlabel=\scriptsize{\textsc{Fraction of Removed Pixels}},
        ]

        \addplot[mark=none, my-blue, thick] 
        table[x=frac,y=sparse_err,col sep=comma] {figures/sparsify_hci/upr.csv};
        
        \addplot[mark=none, my-orange, thick] 
        table[x=frac,y=sparse_err,col sep=comma] {figures/sparsify_hci/ese.csv};
        
        \addplot[mark=none, my-purple, thick] 
        table[x=frac,y=sparse_err,col sep=comma] {figures/sparsify_hci/dpp.csv}; 
        
        \addlegendentry{\scriptsize{UPR}}
        \addlegendentry{\scriptsize{ESE}}
        \addlegendentry{\scriptsize{DPP}}
    \end{axis}
\end{tikzpicture}